\documentclass[10pt,twocolumn,letterpaper]{article}

\usepackage{cvpr}              %

\usepackage{epsfig, graphicx, amsmath, amssymb}
\usepackage{xcolor, tabulary, xfrac, enumitem}
\usepackage[british, english, american]{babel}

\definecolor{citecolor}{RGB}{34,139,34}
\usepackage{nicefrac}
\usepackage{booktabs}
\usepackage{listings}
\usepackage{siunitx}
\usepackage{multirow}
\usepackage[accsupp]{axessibility}  %

\usepackage[pagebackref=true, breaklinks=true, letterpaper=true, colorlinks,
            citecolor=citecolor, bookmarks=false]{hyperref}

\def\cvprPaperID{2588}
\def\confName{CVPR}
\def\confYear{2022}

\renewcommand{\paragraph}[1]{\vspace{1.5mm}\noindent\textbf{#1}}

\newcommand{\tablestyle}[2]{\setlength{\tabcolsep}{#1}\renewcommand{\arraystretch}{#2}\centering\footnotesize}
\newlength\savewidth

\newcommand{\app}{\raise.17ex\hbox{$\scriptstyle\sim$}}

\newcommand{\regnety}{RegNetY\xspace}
\newcommand{\regnetyarch}[1]{RegNetY\,{#1}GF\xspace}

\newcommand{\resnext}{ResNeXt\xspace}

\newcommand{\densenet}{DenseNet\xspace}

\newcommand{\efficientnet}{EfficientNet\xspace}

\newcommand{\vitarch}[1]{ViT {#1}\xspace}

\newcommand{\igdataset}[1][$3.6$B]{IG-#1\xspace}
\newcommand{\indataset}[1]{ImageNet-{#1}k\xspace}

\newcommand{\inatdataset}{iNaturalist\xspace}
\newcommand{\placesdataset}{Places365\xspace}
\newcommand{\placesdatasetfull}{Places365-Standard\xspace}
\newcommand{\cubdataset}{CUB-2011\xspace}
\newcommand{\cubdatasetfull}{Caltech-UCSD Birds-200-2011\xspace}

\newcommand{\superdagger}{\textsuperscript{\textdagger}}

\definecolor{codegreen}{rgb}{0,0.6,0}
\definecolor{codegray}{rgb}{0.5,0.5,0.5}

\begin{document}
\title{Revisiting Weakly Supervised Pre-Training of Visual Perception Models}
\author{%
Mannat Singh  \quad Laura Gustafson \quad Aaron Adcock \quad Vinicius de Freitas Reis \quad Bugra Gedik  \\[2mm]
Raj Prateek Kosaraju \quad Dhruv Mahajan \quad Ross Girshick \quad Piotr Doll\'ar  \ \quad Laurens van der Maaten  \\[2mm]
Meta AI \\[2mm]
{\normalsize \url{https://github.com/facebookresearch/SWAG}}
 }
\maketitle

\begin{abstract}
Model pre-training is a cornerstone of modern visual recognition systems.
Although fully supervised pre-training on datasets like ImageNet is still the de-facto standard, recent studies suggest that large-scale weakly supervised pre-training can outperform fully supervised approaches.
This paper revisits weakly-supervised pre-training of models using hashtag supervision with modern versions of residual networks and the largest-ever dataset of images and corresponding hashtags.
We study the performance of the resulting models in various transfer-learning settings including zero-shot transfer.
We also compare our models with those obtained via large-scale self-supervised learning.
We find our weakly-supervised models to be very competitive across all settings, and find they substantially outperform their self-supervised counterparts.
We also include an investigation into whether our models learned potentially troubling associations or stereotypes.
Overall, our results provide a compelling argument for the use of weakly supervised learning in the development of visual recognition systems.
Our models, Supervised Weakly through hashtAGs (SWAG), are available publicly.
\\

\end{abstract}

\section{Introduction}
\label{sec:introduction}
Most modern visual-recognition systems are based on machine-learning models that are pre-trained to perform a task that is different from the downstream task that the system aims to solve.
Such pre-training allows the system to leverage (annotated) image or video datasets that are much larger than the datasets available for the downstream task.
Arguably the most popular pre-training task is supervised image classification on datasets such as ImageNet and JFT~\cite{dosovitskiy2020image,kornblith2018transfer,zhai2021scaling}, but recent studies have also explored self-supervised~\cite{caron2020unsupervised,caron2021dino,chen2020big,chen2021simsiam,grill2020byol,goyal2021self,kaiming2020moco,misra2020pirl} and weakly supervised~\cite{jia2021align,joulin2016weakly,li2017visual,mahajan2018exploring,radford2021clip} tasks for pre-training.

There are trade-offs between these three types of pre-training.
Fully supervised pre-training benefits from a strong semantic learning signal for each training example, but does not scale well because manual labeling of training data is time-consuming.
By contrast, self-supervised pre-training receives hardly any semantic information on the training examples, but can be scaled to billions of training examples relatively easily~\cite{goyal2021self,kaiming2020moco}.
Weakly-supervised approaches fall somewhere in between: for example, hashtags or other text associated with visual data generally provide a noisy semantic learning signal but can be obtained at large scale with relative ease~\cite{mahajan2018exploring,radford2021clip}.

Following the success of prior work~\cite{mahajan2018exploring}, this paper performs an in-depth study of weakly-supervised pre-training using hashtag supervision.
We pre-train modern image-recognition models on the largest-ever-dataset of images and associated hashtags, and evaluate the resulting models in a range of transfer-learning experiments.
Specifically, we transfer our models to a variety of image-classification tasks and evaluate the performance of the resulting models.
We also evaluate the models in \emph{zero-shot transfer} and \emph{few-shot transfer} settings~\cite{radford2021clip}: that is, we evaluate the ``off-the-shelf performance'' of these models without finetuning them on the target tasks.
The overall goal of our study is to shed light on the trade-offs between fully supervised, self supervised, and weakly supervised pre-training.
Throughout our experiments, we find the weakly-supervised approach to be very competitive: our best models perform on par with the state-of-the-art on a range of visual-perception tasks despite employing a relatively simple training pipeline.

A potential downside of weakly-supervised pre-training is that models may inherit or amplify harmful associations from the underlying supervisory signal.
We perform a series of experiments aimed at assessing the extent to which this happens.
Our results do not provide conclusive answers, but they do suggest that the risks involved may not be as large as in language modeling~\cite{bender2021parrots,gpt3}.
Overall, we believe our study presents a compelling argument for weakly-supervised pre-training of visual-recognition systems.

\section{Related Work}
\label{sec:related_work}
This study is part of a large body of work on pre-training models for visual recognition. 
This body of work can be subdivided into three key groups.

\noindent{\textbf{Fully supervised pre-training}} was pioneered by~\cite{donahue2014,razavian2014} and is now the de-facto standard approach to a variety of visual-recognition tasks, including fine-grained image classification~\cite{he2021transfg,ridnik2021}, object detection~\cite{ren2015faster}, image segmentation~\cite{he2017mask,wang2020maxdeeplab}, image captioning~\cite{li2020oscar}, visual question answering~\cite{kazemi2017vqa}, video classification~\cite{fan2021mvit}, \emph{etc.}
The ImageNet-1K dataset~\cite{russakovsky2015imagenet} is by far the most commonly used image dataset for pre-training, whereas the Kinetics dataset~\cite{kay2017kinetics} is often used for pre-training of video-recognition models.
Some recent studies have also used the much larger JFT-300M~\cite{dosovitskiy2020image} and JFT-3B~\cite{zhai2021scaling} image datasets, but not much is known publicly about those datasets. 
The effectiveness of supervised pre-training has been the subject of a number of studies, in particular,~\cite{abnar2021exploring,kornblith2018transfer,recht2019imagenet} analyze the transfer performance of supervised pre-trained models.

\noindent{\textbf{Self-supervised pre-training}} has seen tremendous progress in recent years.
Whereas early self-supervised learners such as RotNet~\cite{gidaris2018rotation} or DeepCluster~\cite{caron2018deepcluster} substantially lagged their supervised counterparts in vision pre-training, more recent approaches have become quite competitive.
These approaches learn to predict clusters~\cite{caron2020unsupervised}, use contrastive learning~\cite{chen2020big,kaiming2020moco,misra2020pirl}, or use student-teacher architectures in which the teacher is an exponentially moving average of the student~\cite{caron2021dino,chen2021simsiam,grill2020byol}.
A key advantage of self-supervised pre-training is that is can easily be scaled to billions of training images: several studies have shown that scaling self-supervised learning can lead to substantial performance improvements~\cite{goyal2021self,kaiming2020moco}.

\noindent{\textbf{Weakly-supervised pre-training}} has not received nearly as much attention as the other two pre-training paradigms, but has shown very promising performance nonetheless.
Whereas early studies that pre-trained models by predicting words~\cite{joulin2016weakly} or n-grams~\cite{li2017visual} in image captions were not very competitive because of the limited scale of their training data, recent weakly-supervised pre-training methods are much more competitive on a range of visual-recognition tasks~\cite{beal2021billion,ghadiyaram2019large,jia2021align,mahajan2018exploring,radenovic2021large,radford2021clip}.
In particular, ALIGN~\cite{jia2021align} and CLIP~\cite{radford2021clip} pre-train vision-and-language models on large numbers of images and associated captions, and successfully perform \emph{zero-shot transfer} to new recognition tasks.

Our study builds on~\cite{mahajan2018exploring}, which trained convolutional networks on billions of images to predict associated hashtags.
Compared to~\cite{mahajan2018exploring}, our study: (1) trains larger models with more efficient convolutional and transformer architectures on a much larger dataset, (2) studies the performance of the resulting models in zero-shot transfer settings in addition to standard transfer-learning experiments, (3) performs comparisons of our models with state-of-the-art self-supervised learners, and (4) presents an in-depth study of potential harmful associations that models may adopt from the weak supervision they receive.
Despite the conceptual similarities in our approach, our best model achieves an ImageNet-1K validation accuracy that is more than $3\%$ higher than that reported in~\cite{mahajan2018exploring}.

\section{Pre-Training using Hashtag Supervision}
\label{sec:pretraining}
Our weakly supervised pre-training methodology is based on hashtag supervision.
We train image-recognition models to predict the hashtags that were assigned to an image by the person who posted the image.
Hashtag prediction has great potential as a pre-training task because hashtags were assigned to images to make them \emph{searchable}, \ie, they tend to describe some salient semantic aspects of the image.
While hashtag prediction is conceptually similar to image classification, it differs in a few key ways~\cite{denton2015user,mahajan2018exploring,veit2018separating}: 
\begin{enumerate}[topsep=2.5pt, itemsep=-0.8ex]
\item Hashtag supervision is inherently noisy. 
Whilst some hashtags describe visual content in the image (\emph{e.g.}, \texttt{\#cat}), other hashtags may be unrelated to the visual content (\emph{e.g.}, \texttt{\#repost}).
Different hashtags may be used to describe the same visual content, or the same hashtag may be used to describe different visual content.
Importantly, hashtags generally do not provide a comprehensive annotation of the visual content of an image, that is, there tend to be many false negatives.
\item Hashtag usage follows a Zipfian distribution~\cite{mikolov2013distributed}; see Figure \ref{fig:hashtag_histogram}. 
This implies that the learning signal follows a very different distribution than is common in image-recognition datasets like ImageNet~\cite{russakovsky2015imagenet}, which tend to have a class distribution that is more or less uniform.
\item Hashtag supervision is inherently \emph{multi-label}: a single image generally has multiple hashtags associated with it that all serve as positive classification targets.
\end{enumerate}
Our data pre-processing and model pre-training procedures are designed to (partly) address these issues.
We describe them in more detail in Section~\ref{subsec:dataset_collection} and~\ref{subsec:pretraining}, respectively.

\subsection{Hashtag Dataset Collection} 
\label{subsec:dataset_collection}
We follow~\cite{mahajan2018exploring} in constructing a dataset of public Instagram photos and associated hashtags. 
We adopt the following four steps in constructing the pre-training dataset:
\begin{enumerate}[topsep=2.5pt, itemsep=-0.8ex]
	\item Construct a hashtag vocabulary by selecting frequently used hashtags and canonicalizing them.
	\item Gather publicly available images that are tagged with at least one of the selected hashtags.
	\item Combine the resulting images and associated hashtags into labeled examples that can be used for pre-training.
	\item Resample the resulting examples to obtain the desired hashtag distribution.
\end{enumerate}
Next, we describe each of these steps in detail.

\begin{figure}[t]\centering
\includegraphics[width=8.2cm]{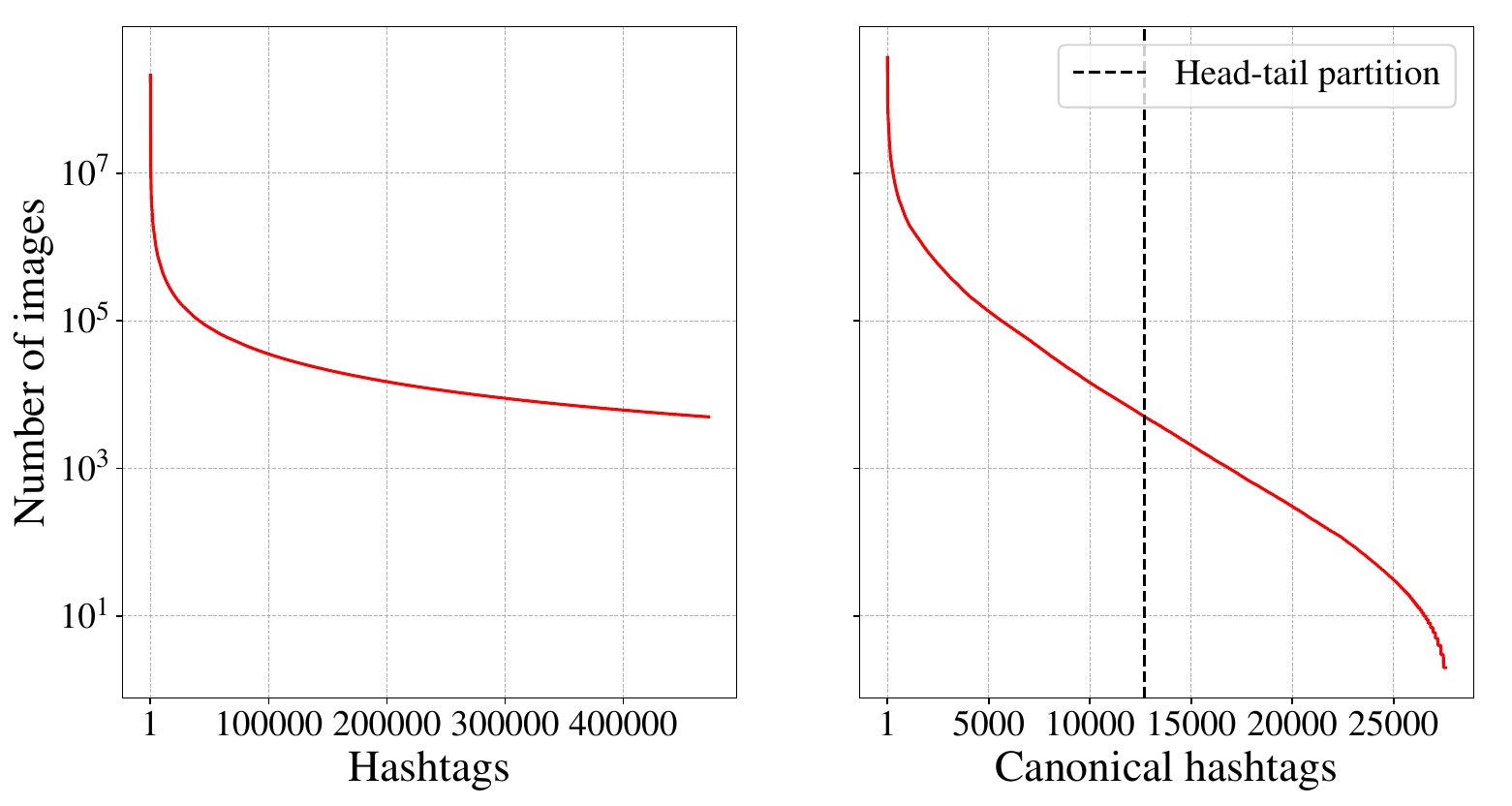}
\caption{Hashtag distribution of Instagram images. \textbf{Left:} Frequency of all hashtags occurring with public images posted by US users. \textbf{Right:} Frequency of filtered and canonicalized hashtags occurring with public images by users from all countries. We define the head as the set of canonical hashtags associated with more than $5{,}000$ images; the remaining hashtags form the tail.}
\label{fig:hashtag_histogram}
\end{figure}

\noindent\textbf{Hashtag vocabulary.} 
\label{subsubsec:hashtag_vocab}
We select hashtags used more than once in public Instagram posts by US users.
Next, we filter out and canonicalize the hashtags using WordNet synsets \cite{fellbaum1998wordnet}. More details about this process are in \iftoggle{cvprfinal}{Appendix \ref{app:hashtag}}{the appendix}.
This results in a label set, $\mathcal{C}$, that contains $\app$27k \emph{canonical hashtags} that correspond to a set of $\app$75k raw hashtags, where multiple hashtags can map to a single canonical hashtag (\emph{e.g.}, \texttt{\#dog} and \texttt{\#canine}). We drop the ``canonical" qualifier when it is obvious from the context.
As the exact images in the dataset may change with time, the number of canonical hashtags varies between 27k and 28k across experiments.
The hashtag selection and canonicalization reduces some of the inherent noise in the supervisory signal.

\noindent\textbf{Image collection and labeling.}
We collect all public Instagram images that have at least one hashtag from our vocabulary.\footnote{We downloaded images from all countries, but excluded images by users from particular countries to comply with applicable regulations.}
The images were subjected to an array of automated filters designed to remove potentially offensive content.
While certainly not perfect, this substantially reduces the issues that plague other large image datasets~\cite{birhane2021multimodal,prabhu2020large}.
We construct a multi-label dataset using these images by converting all hashtags into their corresponding canonical targets (note that a single image may have multiple hashtags). 
Hashtags that are not in the vocabulary are discarded.

\noindent\textbf{Resampling.}
We adopt a resampling procedure similar to~\cite{mahajan2018exploring} to generate our final pre-training examples.
The resampling procedure aims to down-weight frequent hashtags whilst up-weighting infrequent hashtags in the pre-training task.
We do so by resampling according to the inverse square root of the hashtag frequency.
Unlike~\cite{mahajan2018exploring}, we additionally upsample (with replacement) the long tail of images with at least one infrequent hashtag by $\app$100$\times$.
Herein, we define infrequent hashtags as those that occur with fewer than $5{,}000$ images (see Figure \ref{fig:hashtag_histogram}).
The resulting resampled dataset comprises 30\% tail images and 70\% head images (see \iftoggle{cvprfinal}{Appendix~\ref{app:hashtag}}{the appendix} for more details).

We note that this means that in a single training epoch, each unique tail image appears multiple times.
This implies there is a discrepancy between the number of \emph{unique} images in an epoch and the number of \emph{total samples} processed in that epoch. 
We label our dataset by the number of unique images in the dataset: our \igdataset[3.6B] dataset has $\app$3.6 billion unique images.
However, a single training epoch over that dataset processes $\app$5 billion samples due to our re-sampling procedure. 
This is different from other datasets we compare with (\emph{e.g.}, JFT-300M) in which the unique number of images equals the total samples processed in an epoch.

\subsection{Pre-Training Procedure} 
\label{subsec:pretraining}
In preliminary experiments (\iftoggle{cvprfinal}{Appendix~\ref{app:conv_model_family}}{see appendix}), we studied image-recognition models including \resnext~\cite{xie2017aggregated}, \regnety~\cite{radosavovic2020designing}, \densenet~\cite{huang2017densely}, \efficientnet~\cite{tan2019efficientnet}, and ViT~\cite{dosovitskiy2020image} .
We found \regnety and ViT models to be most competitive, and focus on those in the experiments presented here.

During pre-training, we equip our models with an output linear classifier over $|\mathcal{C}| \approx 27$k classes. For ViTs we use an additional linear layer with output dimension equal to the input dimension, similar to \cite{dosovitskiy2020image}.
Following \cite{mahajan2018exploring}, we use a softmax activation and train the model to minimize the cross-entropy between the predicted probabilities and the target distribution. 
Each target entry is either $\nicefrac{1}{K}$ or 0 depending on whether the corresponding hashtag is present or not, where $K$ is the number of hashtags for that image.

All our RegNetY models were trained using stochastic gradient descent (SGD) with Nesterov momentum of $0.9$. 
We employed a half-cosine learning rate schedule~\cite{loshchilov2017sgdr} with a base initial value of 0.1 for a batch size of 256 and a final value of $0$.
We used a weight decay of $10^{-5}$, but disabled weight decay in batch-normalization layers: preliminary experiments suggested that batch-normalization weight decay is effective when pre-training on \indataset{1}, but significantly degrades results on larger datasets such as \igdataset.

Our ViT models were trained using AdamW~\cite{loshchilov2017decoupled} with $\beta_1\!=\!0.9$ and $\beta_2\!=\!0.95$.
We used an initial learning rate of $4 \cdot 10^{-4}$, a batch size of $8{,}192$, and a weight decay of $0.1$.

Following~\cite{goyal2017accurate}, we scale the initial learning rate linearly with the batch size when doing distributed training. 
We ``warm up'' the learning rate for the first $5\%$ of training updates by linearly increasing the learning rate from $\nicefrac{1}{10}$-th of the initial value to the initial value. 
Similar to~\cite{goyal2017accurate}, we find that performance degrades for batch sizes larger than $8{,}192$ so we did not increase our batch size further.

We trained our models using mixed-precision training on images that were pre-processed to $224\!\times\!224$ resolution using a standard random-resize crop followed by a random horizontal flip. %
In preliminary experiments, we also evaluated several other training approaches that provide gains in \indataset{1} pre-training \cite{dollar2021fast,tan2019efficientnet}, including exponential moving averages~\cite{polyak1992}, mixup~\cite{zhang2017mixup}, label smoothing~\cite{muller2019smoothing}, AutoAugment~\cite{cubuk2018autoaugment}, and stochastic depth~\cite{huang2016stochastic}.
However, we did not find those approaches to lead to performance improvements; some even deteriorated performance.

We trained our largest model for 2 epochs of the \igdataset[3.6B] dataset (10 billion samples seen) using 128 Nvidia V100 32GB GPUs across 16 nodes.
The nodes were connected via Ethernet, with 8 GPUs / node connected via NVLink.

\section{Experiments}
\label{sec:experiments}

We performed a series of experiments to test the efficacy of our hashtag-based pre-training strategy.
We compare our weakly supervised models in transfer-learning experiments with modern supervised (Section~\ref{subsec:sota}) and self-supervised models (Section~\ref{subsec:ssl}), and with other weakly supervised models in zero-shot transfer (Section~\ref{subsec:zeroshot}). %

\subsection{Experimental Setup}
\label{subsec:experimental_setup}
In our experiments, we focus on different types of transfer learning to image-classification tasks. 
Specifically, we study: (1) transfer learning using linear classifiers, (2) transfer learning using finetuning, (3) zero-shot transfer learning, and (4) few-shot transfer learning.
We compare the efficacy of our pre-training strategy with that of fully supervised (\ref{subsec:sota}) and self-supervised (\ref{subsec:ssl}) pre-training strategies.

\noindent\textbf{Datasets.}
We perform experiments in which we transfer models to ImageNet classification~\cite{russakovsky2015imagenet} on \indataset{1} (1.28M training images, $50{,}000$ validation images, $1{,}000$ classes), and \indataset{5} ($6.57$M training images, $250{,}000$ validation images, $5{,}000$ classes) as defined in \cite{mahajan2018exploring, xie2017aggregated}.
We also perform experiments in which we transfer pre-trained models to other commonly used image-classification benchmarks, including the iNaturalist 2018~\cite{horn2018inaturalist}, \placesdatasetfull~\cite{zhou2017places}, and \cubdatasetfull (\cubdataset)~\cite{wah2011cub} datasets.

\noindent\textbf{Finetuning.}
We follow~\cite{kolesnikov2019big} in finetuning our pre-training models for downstream tasks.
We finetune the models using SGD with a batch size of 512 and a half-cosine learning rate schedule~\cite{loshchilov2017sgdr}.
The initial value was tuned for every each model-task combination separately via grid-search.
We did not use weight decay during finetuning. 
We finetune RegNetY and ViT B/16 models using an image resolution of $384\!\times\!384$, and ViT L/16 and H/14 models with larger $512\!\times\!512$ and $518\!\times\!518$ resolutions respectively -- higher resolutions help these models significantly. For EfficientNets, we use the pre-training resolution for finetuning.
For ``large'' transfer datasets (defined as datasets with $N\!>\!500{,}000$ examples), we finetune for $20{,}000$ parameter updates;
for ``medium'' datasets ($20{,}000\!<\!N\!\leq\!500{,}000$ examples), we finetune for $10{,}000$ steps; and for ``small'' datasets ($N\!\leq\!20{,}000$ examples), we finetune for 500 steps.
We use mixup~\cite{zhang2017mixup} with $\alpha\!=\!0.1$ during finetuning on all datasets. 
We used synchronous batch normalization across GPUs, as it improves transfer performance (see appendix).

For ImageNet-1k finetuning, we additionally compute an exponential moving average (EMA) of the parameters during training with a decay rate of $10^{-4}$ and use the averaged weights for inference \cite{polyak1992}. We found this improved the top-1 accuracy for our best RegNetY and ViT models by $0.2\%$.  Lastly, we finetuned ViTs for 28 epochs on ImageNet-1k since the longer schedule helped improve performance. 

During evaluation, we resize the smaller side of the image to the final resolution and then take a center crop of the same size (\emph{e.g.}, resize smaller side to 224 then $224\!\times\!224$ center crop). 
This differs from standard practice \cite{touvron2019fixing} but gives a boost of 0.1\% to 0.5\% on the \indataset{1} dataset.

\subsection{Comparison with Supervised Pre-Training}
\label{subsec:sota}
We compare our weakly supervised RegNetY and ViT models with state-of-the-art supervised EfficientNets~\cite{xie2020self,xie2020adversarial} and ViTs~\cite{dosovitskiy2020image,zhai2021scaling} in transfer-learning experiments on five datasets: (1) \indataset{1} , (2) \indataset{5}, (3) \inatdataset , (4) \placesdataset, and (5) \cubdataset. 
We finetune all models (see~\ref{subsec:experimental_setup}) on the training split of the transfer dataset and measure the classification accuracy of the finetuned models on the validation or test split.

\begin{table*}[t]\centering
\resizebox{1.5\columnwidth}{!}{\tablestyle{4pt}{1.05}
\begin{tabular}{@{}lccc|cc|cccc|ccc@{}}
\bf Model & \bf Pre-training & \multicolumn{2}{c|}{\bf Resolution} & \multicolumn{2}{c|}{\bf IN-1k Accuracy} & \multicolumn{4}{c|}{\bf Classification accuracy} & \bf Throughput & \bf FLOPs & \bf Params \\
            &  & Pre. & Fine. & Report. &  Reprod. & \bf IN-5k & \bf iNat. & \bf Places & \bf CUB& (images/sec.) & (B) & (M) \\\midrule
 \multicolumn{7}{l}{\textit{Supervised pre-training}$^\dagger$}\\\midrule
 EfficientNet L2~\cite{xie2020self} & JFT 300M$^\ddagger$ & 475 & 800 & \emph{88.4} & 88.3 & -- & -- & -- & -- & 108 & 479.9 & 480.3 \\
 EfficientNet L2~\cite{xie2020self} & JFT 300M$^\ddagger$ & 475 & --     & \emph{88.2} & 88.0  &  \textbf{61.8} & \textbf{86.5} & 59.4 & 91.2$^\mathsection$ & 293 & 172.6 & 480.3  \\
 EfficientNet B7~\cite{xie2020self} & JFT 300M$^\ddagger$ & 600 & --     & \emph{86.9} & 86.7 & 56.7 & 82.0 & 59.2 & 90.6$^\mathsection$ & 652 & 38.4 & 66.3  \\
 EfficientNet B6~\cite{xie2020self} & JFT 300M$^\ddagger$ & 528 & --     & \emph{86.4} & 86.3 & 55.4 & 79.9 & 58.8 & 89.1$^\mathsection$ & 849 & 19.5 & 43.0 \\
 EfficientNet B8~\cite{xie2020adversarial} & IN-1k                     & 672 & --     & \emph{85.5} & 85.2 & 54.8 & 81.3 & 58.6 & 89.3$^\mathsection$ & 480 & 63.7 & 87.4 \\
 EfficientNet B7~\cite{xie2020adversarial} & IN-1k                     & 600 & --     & \emph{85.2} & 85.0 & 54.4 & 80.6 & 58.7 & 88.9$^\mathsection$ & 652 & 38.4 & 66.3  \\
 EfficientNet B6~\cite{xie2020adversarial} & IN-1k                     & 528 & --     & \emph{84.8} & 84.7 & 53.6 & 79.1 & 58.5 & 88.5$^\mathsection$ & 849 & 19.5 & 43.0 \\
 ViT G/14~\cite{zhai2021scaling} & JFT 3B                & 224 & 518 & \emph{\textbf{90.5}} & -- & -- & -- & -- & -- &  56 & 2826.1 & 1846.3 \\
 ViT L/16~\cite{zhai2021scaling} & JFT 3B              & 224 & 384 & \emph{88.5} & -- & -- & -- & -- & -- &  567 & 191.5 & 304.7 \\
 ViT H/14~\cite{dosovitskiy2020image} & JFT 300M     & 224 & 518 & \emph{\underline{88.6}} & -- &  -- & -- & -- & -- & 116  & 1018.8 & 633.5 \\
 ViT L/16~\cite{dosovitskiy2020image} & JFT 300M      & 224 & 512 & \emph{87.8} & -- & -- & -- & -- & -- &  255 & 362.9 & 305.2 \\
 ViT L/16~\cite{dosovitskiy2020image} & IN-21k           & 224 & 384 & \emph{85.2} & 85.2 & --      & 81.7 & 59.0 & 91.3$^\mathsection$ & 567 & 191.5 & 304.7 \\
 ViT B/16~\cite{dosovitskiy2020image} & IN-21k           & 224 & 384 & \emph{84.0} & 84.2 & --        & 79.8 & 58.2 & 90.8$^\mathsection$ & 1,161 & 55.6 & 86.9 \\
 ViT L/32~\cite{dosovitskiy2020image} & IN-21k           & 224 & 384 & \emph{81.3} & 81.5 & --       & 74.6 & 57.7 & 88.7$^\mathsection$ & 1,439 & 54.4 & 306.6 \\\midrule
 \multicolumn{7}{l}{\textit{Weakly supervised pre-training}}\\\midrule
 ViT H/14             & IG 3.6B                      & 224 & 518 & \multicolumn{2}{c|}{\underline{88.6}} &  \underline{60.9} & \underline{86.0} & \textbf{60.7} & \textbf{91.7} & 116 & 1018.8 & 633.5\\
 ViT L/16             & IG 3.6B                      & 224 & 512 & \multicolumn{2}{c|}{88.1} &  59.0 & 84.2 & \textbf{60.7} & \underline{91.6} & 255 & 362.9 & 305.2 \\
 ViT B/16             & IG 3.6B                      & 224 & 384 & \multicolumn{2}{c|}{85.3} &  54.5 & 79.9 & 59.1 & 89.8 & 1,161 & 55.6 & 86.9 \\
 RegNetY 128GF & IG 3.6B                      & 224 & 384 & \multicolumn{2}{c|}{88.2} & \underline{60.9} & 85.7 & 60.1 & 90.8 & 307 & 375.2 & 644.8 \\
 RegNetY 32GF & IG 3.6B                      & 224 & 384 & \multicolumn{2}{c|}{86.8} & 58.5 & 82.9 & 59.6 & 89.5 & 976 & 95.1 & 145.0 \\
 RegNetY 16GF & IG 3.6B                      & 224 & 384 & \multicolumn{2}{c|}{86.0} & 57.2 & 81.4 & 59.2 & 88.3 & 1,401 & 47.0 & 83.6 \\
\end{tabular}}
\vspace{2mm}
\caption{Transfer-learning accuracy of models pre-trained on the specified pre-training dataset followed by finetuning and testing on five transfer datasets. Accuracies that were adopted from the original papers are \emph{italicized}. The best result on each dataset is \textbf{boldfaced}; the second-best result is \underline{underlined}. Our weakly-supervised pre-trained models achieve the best or second-best performance on all five transfer datasets. $^\dagger$It is unknown how much manual curation was performed to annotate the JFT datasets. $^\ddagger$IN-1k is used as supervised pre-training data; JFT 300M is used without labels. $^\mathsection$Model was pre-trained on IN-1k training set, which overlaps with the \cubdataset test set.}
\label{tab:transfer_results}
\end{table*}

Table~\ref{tab:transfer_results} presents an overview of the results of these experiments.
For each model, the table shows the pre-training dataset used, the image resolution used during pre-training and finetuning, the inference throughput of the model, the number of FLOPs and parameters in the finetuned model, and the test accuracy on the transfer datasets.
We do not report results for an approach when its pre-trained model and pre-training dataset are not publicly available.
In the table, accuracies that we adopted from the original paper are \emph{italicized}.
For the \indataset{1} dataset, we report both results reported in the original papers and results we obtained when we reproduced the model.
We \textbf{boldface} the best result and \underline{underline} the second-best result for each dataset.
Table~\ref{tab:transfer_results} groups models into supervised and weakly supervised.
In this grouping, we consider pre-training on JFT datasets to be supervised pre-training but we acknowledge that little is known on how these datasets were collected: \cite{zhai2021scaling} refers to the JFT-3B dataset as ``weakly labeled'' and ``noisy'' but also states that semi-automatic annotation was used to collect it.
This suggests that JFT datasets were manually curated and annotated, which is why we consider them as supervised.\footnote{Although our system-level evaluations hamper exact comparisons, our results suggest that the weakly supervised  \igdataset[3.6B] dataset provides the same amount of supervisory signal as the supervised JFT-300M dataset.} 

The results in Table~\ref{tab:transfer_results} show that our weakly-supervised models are very competitive: they achieve the best or second-best accuracy on all five transfer datasets. 
We note that models pre-trained on IN-1k datasets observe $5\%$ of the CUB test data during pre-training~\cite{mahajan2018exploring} as a result of which their performance is overestimated.
This makes the strong performance of our weakly-supervised models (which do not see test data during training) particularly noteworthy.

\begin{figure}[t]\centering
\includegraphics[width=\linewidth]{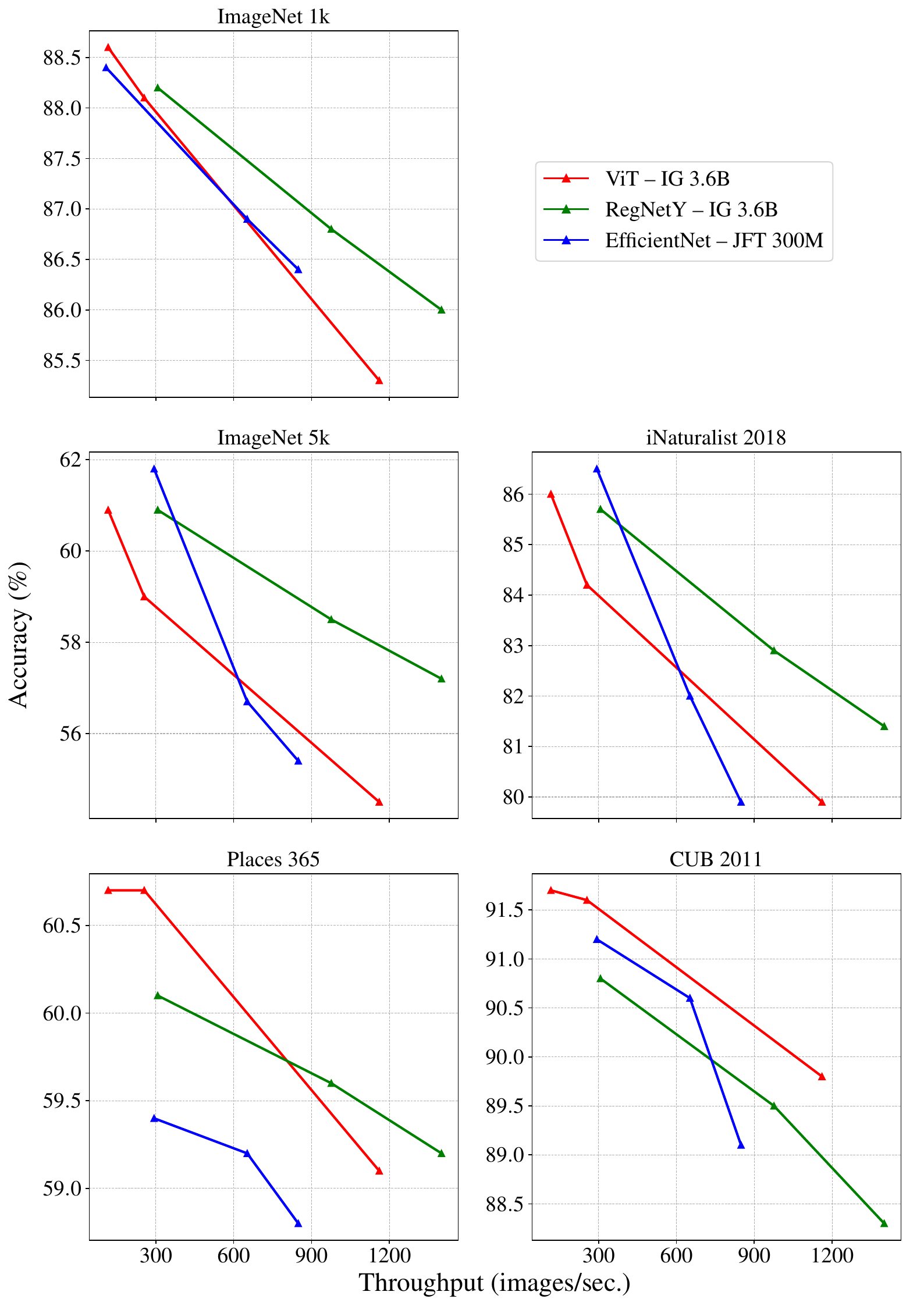}
\caption{Transfer-learning accuracy as a function of throughput of pre-trained models that were finetuned on five datasets (please refer to Table \ref{tab:transfer_results} for full results). ViTs and EfficientNets achieve the highest top-line accuracies, but RegNetY models perform better in the high-throughput regime.}
\label{fig:transfer_throughput}
\end{figure}

To provide more insight into the classification accuracy and throughput trade-off, we plot one as a function of the other in Figure~\ref{fig:transfer_throughput}.
Comparing ViT and RegNetY models trained on the same \igdataset dataset, we observe that vision transformers obtain the highest classification accuracies.
In terms of accuracy-throughput tradeoff, RegNetYs outperform at small to medium model sizes.
The RegNetY 128GF model performs quite similarly on accuracy and throughput to the semi-supervised EfficientNet L2 model, but at smaller size scales, RegNetYs provide a better tradeoff.

\subsection{Comparison with Self-Supervised Pre-Training} 
\label{subsec:ssl}
Our experiments so far suggest that the ability to scale up weakly-supervised pretraining to billions of images 
can offset the lower amount of learning signal obtained per training example.
This raises the question if we need weak supervision at all, or whether modern \emph{self-supervised} learners~\cite{caron2018deepcluster,caron2020unsupervised,caron2021dino,chen2020big,chen2021simsiam,gidaris2018rotation,goyal2021self,grill2020byol,kaiming2020moco,misra2020pirl} may suffice.
Self-supervised learning scales even more easily than weakly-supervised learning, and prior work has demonstrated the potential of self-supervised pre-training at scale~\cite{goyal2021self,kaiming2020moco}.

We perform transfer-learning experiments on ImageNet-1k that compare our weakly-supervised learner with SimCLR v2~\cite{chen2020big}, SEER~\cite{goyal2021self}, and BEiT~\cite{bao2021beit}.
The comparison with SEER is of particular interest: because it is trained on a similar collection\footnote{The data distribution used in~\cite{goyal2021self} and in our study may not be exactly the same, as we use the data resampling approach described in Section~\ref{subsec:dataset_collection}.} of Instagram images, we can readily compare both learning paradigms on the same data distribution.
We perform experiments in two transfer-learning settings: (1) a setting in which a linear classifier is attached on top of the pre-trained model and the resulting full model is finetuned and (2) a setting that initializes this linear classifier using the zero-shot transfer approach described in Section~\ref{subsec:zeroshot} (without Platt scaling) before finetuning the full model.
Following prior work~\cite{chen2020big,goyal2021self}, we vary the amount of labeled ImageNet examples used for finetuning to 1\%, 10\%, and 100\% of the original ImageNet-1k training set.
We report results using images of size $224\!\times\!224$ pixels.

The results of our experiments are presented in Table~\ref{tab:ssl_lowshot:en}.
Results for SimCLRv2, SEER and BEiT were adopted from~\cite{chen2020big,goyal2021self,bao2021beit}; small differences in experimental setup may exist.
Our results show that weakly-supervised learning substantially outperforms current self-supervised learners, particularly in low-shot transfer settings, likely because our weakly-supervised learners %
receive more learning signal per sample.
Moreover, our results show that weakly-supervised learners benefit from zero-shot initialization in low-shot transfer settings.
We note that our observations may change if self-supervised learners are scaled further. %

\subsection{Zero-Shot Transfer}
\label{subsec:zeroshot}
Another potential advantage of weakly-supervised models is that they have observed a large variety of training targets during pre-training.
This may help them recognize new visual concepts quickly.
We test the ability of our models to learn and recognize new visual concepts rapidly in \emph{zero-shot transfer} learning setting.\footnote{Some prior work refers to this learning setting as zero-shot learning~\cite{jia2021align,radford2021clip}. We find this term confusing because it differs from classical zero-shot learning~\cite{lampert2009}. Hence, we adopt the term \emph{zero-shot transfer}.}
In this setting, we use the output layer of the pre-trained model directly without any finetuning. 
We can do this because we trained on 27k hashtags derived from WordNet \cite{fellbaum1998wordnet}, allowing us to define a mapping between hashtags and class labels for datasets, like \indataset{1}, also built on WordNet.
We use the same image resolution as pre-training, \emph{viz.}, $224\times224$ pixels.

\begin{table}[t] \centering
\resizebox{\columnwidth}{!}{\tablestyle{4pt}{1}
\begin{tabular}{@{}lccc|ccc@{}}
\bf Model  & \bf Approach & \bf Pre-training & \bf Transfer &  \multicolumn{3}{c}{\bf Accuracy}  \\
 & & & & \bf 1\% & \bf 10\% & \bf 100\%  \\\midrule
\multicolumn{7}{l}{\textit{Self-supervised pre-training}}\\\midrule
RN152w3 + SK & SimCLRv2$^\dagger$~\cite{chen2020big} & IN-1k & Finetune & \emph{74.9} & \emph{80.1} & \emph{83.1} \\
RegNetY 128GF & SEER~\cite{goyal2021self} & IG 1B & Finetune & \emph{57.5} & \emph{76.7} & \emph{83.8} \\
RegNetY 256GF & SEER~\cite{goyal2021self} & IG 1B  & Finetune & \emph{60.5} & \emph{77.9} & \emph{84.2} \\
ViT L/16 & BEiT~\cite{bao2021beit} & IN-1k  & Finetune & -- & -- & \emph{85.2} \\\midrule
\multicolumn{7}{l}{\textit{Weakly supervised pre-training}}\\\midrule
RegNetY 128GF & Ours & IG 3.6B & ZS-Init.+Ft. & \textbf{82.0} & \textbf{84.5} & \underline{87.8} \\
RegNetY 32GF & Ours & IG 3.6B & ZS-Init.+Ft. &  \underline{79.4} & 82.0 & 86.5 \\
RegNetY 16GF & Ours & IG 3.6B & ZS-Init.+Ft. & 77.6 & 80.8 & 85.7 \\
RegNetY 128GF & Ours & IG 3.6B & Finetune & 79.2 & \underline{84.1} & \textbf{87.9}\\
RegNetY 32GF & Ours & IG 3.6B & Finetune & 74.8 & 81.7 & 86.3 \\
RegNetY 16GF & Ours & IG 3.6B & Finetune & 72.3 & 80.4 & 85.3 \\
\end{tabular}}
\vspace{2mm}
\caption{Transfer accuracy of models on the ImageNet-1k dataset as a function of the percentage of ImageNet-1k training examples used for transfer learning. Transfer learning is performed using either standard finetuning, or zero-shot (ZS) transfer initialization followed by finetuning. The best result in each setting is \textbf{boldfaced}; the second-best result is \underline{underlined}. Accuracies that are adopted from the original paper are \emph{emphasized}. Our weakly supervised pre-trained models outperform models pre-trained with modern self-supervised learners, in particular, in the few-shot regime. $^\dagger$During finetuning, SimCLRv2 accessed 100\% of the ImageNet training images but $k\%$ of the labels, whereas SEER and our method accessed $k\%$ of the training data.
}
\label{tab:ssl_lowshot:en}
\end{table}

\paragraph{Platt scaling.} In our zero-shot transfer experiments, we consider a transductive learning setting~\cite{gammerman1998} in which all test examples are available simultaneously at test time.
This allows us to train a Platt scaler \cite{platt1999} on the test data that corrects for differences in the distribution of hashtags (which are Zipfian) and the distribution of classes in the target task (which is uniform).
The Platt scaler is parameterized by a weight vector $\mathbf{w} \in \mathbb{R}^C$ and bias vector $\mathbf{b} \in \mathbb{R}^C$, where $C$ is the number of classes.
Given a probability vector $\mathbf{p} \in \Delta_C$ with $\Delta_C$ the $C$-simplex, the Platt scaler computes a new output $\mathbf{p}' = \textrm{softmax}\left(\textrm{diag}(\mathbf{w}) \mathbf{p} + \mathbf{b}\right)$.
The Platt scaler is trained to minimize the cross-entropy loss between the test distribution of $\mathbf{p}'$ and a uniform distribution over the $C$ classes.
Note that this does not use the test labels; it only encourages the predictions to be uniform over classes.

\paragraph{Mapping from hashtags to ImageNet classes.}
Because the targets in both the ImageNet and \igdataset datasets are English nouns, we can construct a many-to-many mapping between Instagram hashtags and ImageNet classes.
To do so, we first map both hashtags and ImageNet classes to WordNet synsets, and then map hashtags to ImageNet classes based on their similarity in WordNet \cite{fellbaum1998wordnet}.
We use the resulting many-to-many mapping between hashtags and classes to aggregate hashtag-prediction scores over ImageNet classes.
We experiment with three different aggregation methods and use the method that we found to work best for each model; see appendix for details.

\paragraph{Results.}
The results of our zero-transfer results are presented in Table~\ref{tab:zeroshot:en}.
The table presents top-1 classification accuracies on four ImageNet-like test sets for our models with and without Platt scaling.
We compare the performance of our models with that of CLIP~\cite{radford2021clip} and ALIGN~\cite{jia2021align}.
These experiments are \emph{system-level} comparisons in which many factors are different: 
For example, CLIP was trained on a dataset of 400 million images and captions that appears more curated than ours, it was finetuned at a higher resolution, and it performs zero-shot transfer via prompt engineering~\cite{gpt3} which is known to improve recognition accuracy~\cite{radford2021clip}.
ALIGN uses a different image-recognition model (\emph{viz.}, EfficientNet) and was trained on 1 billion pairs of web images and corresponding alt-texts~\cite{jia2021align}.

Table~\ref{tab:zeroshot:en} presents our results with zero-shot transfer on four ImageNet-like datasets.
The results show that our weakly supervised models perform very well out-of-the-box: without ever seeing an ImageNet image, our best model achieves an ImageNet top-1 accuracy of $75.3\%$.
The results also show that Platt scaling is essential to obtain good zero-shot transfer performance with our model, as it corrects for differences in the distribution of hashtags and ImageNet classes.
Finally, we find that our ViT models underperform our RegNetY models in the zero-shot transfer setting. 
This is unsurprising considering that ViTs also underperformed RegNetYs on ImageNet-1k finetuning at an image resolution of $224\times224$ pixels.

Comparing our models with CLIP~\cite{radford2021clip}, we observe that the CLIP ViT L/14 model slightly outperforms our model in zero-shot transfer to the IN-1k dataset; whereas the smaller RN50$\times$64 CLIP model underperforms it.
On some datasets, the ALIGN~\cite{jia2021align} model performs even slightly better.
However, the results are not fully consistent: our models do obtain the best performance on the ImageNet-v2 dataset~\cite{recht2019imagenet}.
Because these experiments perform system-level comparisons, it is difficult to articulate what drives these differences in performance.
Nonetheless, our results provide further evidence that weakly-supervised approaches like ours, CLIP, and ALIGN provide a promising path towards the development of open-world visual-recognition models~\cite{horn2017devil}.

\begin{table}[t]\centering
\resizebox{0.8\columnwidth}{!}{\tablestyle{4pt}{1}
\begin{tabular}{@{}lc|cccc@{}}
\bf Model                   & \bf Platt &  \multicolumn{4}{c}{\bf Classification accuracy}  \\
                             &                      & \bf IN-1k & \bf ReaL-IN & \bf IN-v2 & \bf Obj. Net  \\\midrule
 Visual n-grams~\cite{li2017learning} & N/A & \emph{35.2} & -- & -- & -- \\
 CLIP RN50$\times$64~\cite{radford2021clip} & N/A & \emph{73.6} & -- & -- & -- \\
 CLIP ViT L/14~\cite{radford2021clip} & N/A & \underline{\emph{76.2}} & -- & \underline{\emph{70.1}} & \textbf{\emph{72.3}} \\
 ALIGN~\cite{jia2021align} & N/A & \textbf{\emph{76.4}} & -- & \underline{\emph{70.1}} & -- \\\midrule
 RegNetY 128GF  & Yes                 & 75.3                 &        \textbf{79.5}  & \textbf{71.1} & \underline{64.3} \\
 RegNetY 32GF  & Yes                 & 73.6                 &        \underline{78.3}  & 69.1 & 49.9 \\
 RegNetY 16GF  & Yes                 & 72.5                 &        77.6  & 67.9 & 45.1 \\
 RegNetY 128GF  & No                 &  65.1                  &        69.7                  & 60.2                     & 54.2 \\
 RegNetY 32GF  & No                 &  62.2                &        67.5  & 57.3 & 59.1 \\
 RegNetY 16GF  & No                 &   60.7               &        66.3  & 55.6 & 54.8 \\
 ViT H/14  & Yes                 & 72.3                &       76.5  &  66.5 & 60.0 \\
 ViT L/16  & Yes                 & 71.6                 &        76.0  & 65.7 & 57.3 \\
 ViT B/16 & Yes                 &  67.7                  &       73.0                 & 61.9                    & 43.0 \\
 ViT H/14 & No                 &  62.8                  &        67.3                  & 57.7                     & 52.4 \\
 ViT L/16  & No                 &  62.1               &        66.6  & 56.3 & 51.1 \\
 ViT B/16  & No                 &   58.4               &        63.6  & 52.3 & 48.9 \\
\end{tabular}}
\vspace{2mm}
\caption{Zero-shot transfer accuracy of models on four datasets with WordNet-based classes: (1) the ImageNet-1k dataset, (2) the ReaL ImageNet~\cite{beyer2020imagenet} dataset, (3) the ImageNet v2~\cite{recht2019imagenet} dataset, and (4) the ObjectNet~\cite{barbu2019objectnet} dataset. The best result on each dataset is \textbf{boldfaced}; the second-best result is \underline{underlined}. Accuracies that are adopted from the original paper are \emph{italicized}. When using Platt scaling, our weakly-supervised RegNetY models work very well out-of-the-box. They achieve $75.3\%$ zero-shot transfer accuracy on ImageNet-1k, and outperform CLIP~\cite{radford2021clip} and ALIGN~\cite{jia2021align} on the ImageNet v2~\cite{recht2019imagenet} dataset.}
\label{tab:zeroshot:en}
\end{table}

\section{Broader Impact}
\label{sec:broader_impact_summary}
A potential downside of weakly-supervised training of models on uncurated web data is that they may learn harmful associations that reflect offensive stereotypes~\cite{bender2021parrots,gpt3}.
Moreover, the models may not work equally well for different user groups; for example, they do not work as well in non-English speaking countries~\cite{devries2019} because we used English hashtags as the basis for training our models.
We performed a series of experiments to better understand: (1) the associations our hashtag-prediction models learn with photos of people with varying characteristics, and (2) how well those models perform on photos taken in non-English speaking countries.
We summarize the results of those experiments here and refer to the appendix for further details.

\noindent\textbf{Analyzing associations in hashtag predictions.}
We performed experiments analyzing the associations our RegNetY 128GF hashtag-prediction models make for photos that contain people with different apparent skin tone, apparent age, apparent gender, and apparent race.
The experiments were performed using: (1) a proprietary dataset that contains $178{,}448$ Instagram photos that were annotated using the Fitzpatrick skin tone scale~\cite{fitzpatrick1975} and (2) the UTK Faces dataset, which provides apparent age, apparent gender, and apparent race labels~\cite{zhifei2017utk}.

We find that the model has learned several associations between hashtags and skin tone; see the appendix for details.
For example, \texttt{\#redhead} is more commonly predicted for photos of people with a light skin tone, whereas \texttt{\#black} is more often predicted for people with a dark skin tone. 
Similarly, some hashtag predictions correlate with the apparent age of people in photos; see the appendix for details.
For example, our models more commonly predict \texttt{\#baby} or \texttt{\#kid} for photos that contain people who are $1\!-\!10$ years old, and more commonly predict \texttt{\#elder} for the $80\!-\!90$ years age group.
When analyzing our model for gender stereotypes, we found that our model's hashtag predictions associate men with \texttt{\#football} and \texttt{\#basketball} more frequently.
By contrast, our model associates photos containing women more frequently with \texttt{\#makeup} and \texttt{\#bikini}; see the appendix for details.

The most troubling associations we observed stem from an analysis of model predictions for photos that contain people with different apparent race.
In particular, some of our experiments suggest that our model may associate photos that contain Black people with \texttt{\#mugshot} and \texttt{\#prison} more frequently; see the appendix.
However, it is unclear whether these observations are due to our model making incorrect or biased predictions for photos in the evaluation dataset, or whether they are due to the evaluation dataset containing a problematically biased image distribution.
In particular, a more detailed analysis uncovered the presence of a troubling bias in the evaluation dataset (rather than in our model): we found that the UTK Faces dataset~\cite{zhifei2017utk} contains a substantial number of mug shots that disproportionally portray Black individuals.

Overall, our results suggest that while our hashtag-prediction models appear to make fewer troubling predictions than language models~\cite{bender2021parrots,gpt3}, careful analyses and adaptations would be needed before hashtag predictions from our model can be used in real-world scenarios.
Motivated by this observation, we do not release the final hashtag-prediction layer of our models as part of this study.

\noindent\textbf{Analyzing hashtag prediction fairness.}
We also analyzed how well our hashtag-prediction models work on photos taken across the world. 
We repeated the analysis of~\cite{devries2019} on the Dollar Street dataset and performed analyses on a proprietary dataset that contains millions of images with known country of origin.
Akin to~\cite{devries2019}, we observe large accuracy differences of our model on Dollar Street photos from different countries.
Our analysis on the much larger and more carefully collected proprietary dataset confirms this result but suggests that the effect sizes are much smaller than reported in~\cite{devries2019}; see the appendix for details.
Specifically, we find that the range of per-country accuracies is in a relatively tight range of $\app5\%$ \ie, our model achieves per-country recognition accuracies between $65\%$ and $70\%$ for all $15$ countries in the dataset.
Overall, our results suggest more work is needed to train models that perform equally across the world.
In future work, we plan to train multi-lingual hashtag models~\cite{sigurdsson2020grounding} as this may lead to models that achieve equal recognition accuracies across countries.
\section{Discussion}
In this paper, we have presented an in-depth study of fully supervised, self-supervised, and weakly-supervised pre-training for image recognition.
Combined with related work~\cite{ghadiyaram2019large,jia2021align,mahajan2018exploring,radenovic2021large,radford2021clip}, our results provide a compelling argument for the use of weakly-supervised pre-training in the development of systems for visual perception.
However, our study also uncovers limitations of this line of research.

In particular, we find it is increasingly difficult to perform systematic, controlled experiments comparing different approaches and techniques.
There are a variety of reasons for this, including the use of proprietary data that was collected via opaque processes\footnote{We acknowledge that, although the data we use in our experiments is public, it is hard for others to collect that data. However, unlike other studies, we did strive to be comprehensive in describing our data-collection procedure, as we aim to maximize what the reader can learn from our study.}, the diversity of model architectures used, the complexity of training recipes, the heterogeneity of hardware and software platforms used, the vast compute resources required, and the fact that not all studies publish pre-trained models.
Together, this creates an environment in which researchers cannot perform controlled studies that test the effect of one variable, keeping all other variables fixed. 
Instead, they can only perform \emph{system-level} comparisons, as we did in this study.
Such comparisons provide signal on the potential of various approaches, but they do not produce conclusive results.
This problem is exacerbated by the fact that the signal we are measuring is small, as recognition accuracies on commonly used evaluation datasets appear saturated.
To create a thriving research community focused on large-scale learning of vision systems, it is imperative that we address these issues.

A second limitation of this line of research is the strong focus on recognition accuracy and inference speed as the main measures of merit.
While recognition accuracy and inference speed are obviously important, they are not the only measures that matter for the quality of a visual-perception system.
Other measures include the recognition accuracy experienced by different groups of users and the prevalence of predictions that reinforce harmful stereotypes.
We presented an initial study of such measures in Section~\ref{sec:broader_impact_summary} but this foray is not completely conclusive or sufficient.
In particular, we found there are no well-established evaluation datasets and experimental protocols that facilitate the rigorous analyses.
To make matters worse, the presence of harmful stereotypes in some commonly used vision datasets (such as the association between Black people and mug shots we found in the UTK Faces dataset~\cite{zhifei2017utk}) appears to be unknown.
In order to make hashtag-prediction systems like ours ready for real-world deployment, it is essential that we improve the quality of our analyses, and that we address any issues that those analyses may surface.

To conclude, we emphasize that we remain convinced about the potential of weakly-supervised learning approaches.
If we resolve the aforementioned issues, we believe such approaches may improve visual-perception systems in the same way that large-scale language models have improved natural language understanding, machine translation, and speech recognition.

\iftoggle{cvprfinal}{
\section*{Acknowledgements}
We would like to thank Ishan Misra, Priya Goyal, Benjamin Lefaudeux, Min Xu and Vinayak Tantia for discussions and feedback, and Haowei Lu and Yingxin Kang for help with the data loader implementation.
We thank Deepti Ghadiyaram, Anmol Kalia, and Katayoun Zand for their work on the internal datasets with apparent skin tone and country annotations.
We thank Phoebe Helander, Adina Williams, Maximilian Nickel and Emily Dinan for helpful feedback on the Broader Impact analysis.
Lastly, we thank Brian O'Horo for support with the training infrastructure. 
}{}

{\small\bibliographystyle{ieee_fullname} \bibliography{refs}}

\iftoggle{cvprfinal}{
\clearpage
\section*{Appendix}
\appendix
\section{Additional Dataset Details}
\label{app:hashtag}

\noindent\textbf{Hashtag Filtering and Canonicalization.} We considered the set of all hashtags $\mathcal{H}$ posted by US users more than once in public posts as our candidate set. We design a many-to-one function to map a hashtag to WordNet synsets \cite{fellbaum1998wordnet}, $s: \mathcal{H} \rightarrow 2^\mathcal{S}$, where $\mathcal{S}$ is the set of WordNet synsets, and $2^\mathcal{S}$ is the power set of $\mathcal{S}$. $s$ is defined as the \texttt{get\_synsets()} Python function in Listing \ref{lst:hashtag_synset}. We filter out hashtags which map to $\emptyset$, and consider all hashtags which map to the same set of synsets as the same label. For instance, \texttt{\#eggplant} and \texttt{\#aubergine} map to the same target, whereas \texttt{\#newyork} is filtered out. We finally convert the output of $f(h)$, a set of synsets, to a text string, which we refer to as a \emph{canonical hashtag}. We refer to the set of all canonical hashtags, which is our output vocabulary, by $C$.

\begin{minipage}{0.5\textwidth}
\lstdefinestyle{codestyle}{
    keywordstyle=\color{magenta},
    numberstyle=\tiny\color{codegray},
    stringstyle=\color{codegreen},
    basicstyle=\ttfamily\scriptsize,
    breakatwhitespace=false,         
    breaklines=true,                 
    captionpos=b,                    
    keepspaces=true,                 
    numbers=left,                    
    numbersep=5pt,                  
    showspaces=false,                
    showstringspaces=false,
    showtabs=false,                  
    tabsize=2
}

\lstset{style=codestyle}

\begin{lstlisting}[language=Python, label={lst:hashtag_synset}, caption={Hashtag-to-synset mapping code in Python.~~~~~~~~~~~}]
from nltk.corpus import wordnet

MIN_LEN = 3
ALLOWED_SENSES = {
  "noun.animal",
  "noun.artifact",
  "noun.food",
  "noun.object",
  "noun.plant",
  "noun.event",
}


def get_synsets(hashtag):
  if len(hashtag) < MIN_LEN:
    return set()

  candidates = {wordnet.morphy(hashtag, wordnet.NOUN)}
  for i in range(MIN_LEN, len(hashtag) - MIN_LEN + 1):
    candidate = hashtag[:i] + "_" + hashtag[i:]
    candidates.add(wordnet.morphy(candidate))

  synsets = set()
  for word in candidates:
    if word is None:
      continue
    for synset in wordnet.synsets(word):
      if synset._lexname in ALLOWED_SENSES:
        synsets.add(synset)
  return synsets

\end{lstlisting}
\end{minipage}

\noindent\textbf{Image Sampling.}
We follow \cite{mahajan2018exploring} and down-weight the relative weight of frequent hashtags. %
For deciding our sampling weights for images, we assign importance factors to each image based on the (canonical) hashtags associated with it. For a hashtag $h \in \mathcal{C}$, its importance factor, $I_h$, is defined as $f(h)^{-1/2}$, where $f(h)$ is the hashtag's frequency. For an image $i$, with associated hashtags $\{h_i^j\}$, we define the image's importance factor as $I_i = \max{I_{h_i^j}}$.
Next, we partition the hashtags into two sets -- the head, which contains hashtags which occur in at least $5000$ images, and the tail which contains the remaining infrequent hashtags\iftoggle{cvprfinal}{ (see Figure \ref{fig:hashtag_histogram})}{}. An image is considered a tail image iff it contains at least one tail hashtag. 

We sample images from the set of all images available to us, $\mathcal{I}$, using the probability distribution $p_i = c I_i  \; \forall i \in \mathcal{I}$, where $c$ is a normalization constant. We continue sampling images independently until we reach our target dataset's total samples. For a target number of samples, $M$, we sample $\alpha M$ samples from the head and $(1 - \alpha) M$ samples from the tail using this sampling procedure (we chose $\alpha = 0.7$). We note that because the tail is heavily upsampled, the number of unique images in a single epoch is smaller than the total samples $M$.

\begin{table}[t]\centering
  \resizebox{0.8\columnwidth}{!}{
  \tablestyle{4pt}{1}
  \begin{tabular}{@{}l|c|c|c@{}}
  \bf Approach & \bf Hashtag & \bf Head-tail & \bf IN-1k \\
    & \bf vocabulary & \bf sampling & \bf Accuracy \\
    & \bf size & \bf ratio ($\boldsymbol{\alpha}$) & \\\midrule
    Baseline \cite{mahajan2018exploring} & 17K & - & 74.9 \\\midrule
    \multirow{3}{*}{+ Head-tail sampling} & 17K & 0.7 & 76.6 \\  
     & 17K & 0.5 & 76.3 \\  
     & 17K & 0.3 & 75.5 \\\midrule
    + Larger hashtag vocab. & 27K & 0.7 & 77.0 \\
  \end{tabular}
  }
  \vspace{2mm}
  \caption{\textbf{Dataset ablations}. 
  Ablation study on training set collection using ResNeXt-101 32x8d models trained on IG datasets with 100M unique images; 
  we report the linear classifier accuracy on \indataset{1}.
  The baseline approach follows the dataset collection approaches in \cite{mahajan2018exploring} and reproduces the results in that paper.
  Partitioning the hashtags and over-sampling the tail ($\alpha\!=\!0.7$) improves transfer accuracy significantly,
  but excessively over-sampling the tail ($\alpha\!=\!0.3$) worsens it. Increasing the hashtag vocabulary size improves transfer accuracy.}
  \label{tab:dataset_ablations}
  \end{table}

The deviations from \cite{mahajan2018exploring} in dataset collection were ablated by pre-training on datasets of 100 million samples and evaluating linear classifier performance on ImageNet-1k, see Table \ref{tab:dataset_ablations} for details.
Per the results in the table, our changes boost transfer performance on ImageNet-1K by $2.1\%$ when pretraining a ResNeXt-101 32x8d on a 100M dataset.
This number might change as we increase the size of the dataset or model.

\textbf{Deduplication.}
\cite{mahajan2018exploring} performed an extensive deduplication experiment, which suggests that the percentage of images in common evaluation datasets that appears on Instagram is very small ($<$ 0.5\%) and, in fact, smaller than the overlap between those evaluation datasets and the ImageNet training set that is commonly used for model pre-training.
While our sampling methodologies may differ, based on those observations, we chose not to repeat the deduplication experiments.

\section{Model Complexity and Speed}
\label{app:model_complexity}

Table~\ref{tab:model_complexity} presents the resolution, FLOPs, number of parameters, number of activations, and train and test throughputs of all models used in our study.

\begin{table}[!htb]\centering

\resizebox{\columnwidth}{!}{\tablestyle{4pt}{1}
\begin{tabular}{@{}c|c|ccc|cccc@{}}
 \bf Model & \bf Resolution    & \bf Flops  & \bf Params & \bf  Acts & \bf Train & \bf Test\\
           &                       &  (B)      &    (M)   &  (M)  &  (images/sec.) &  (images/sec.) \\\midrule
 EfficientNet L2 & 475  & 172.6 & 480.3 & 609.9 & 49      & 293 \\
 EfficientNet L2 & 800  &  479.9 & 480.3 & 1707.4 & 19  & 108 \\
 EfficientNet B8 & 672  & 63.7 & 87.4 & 442.9    & 103      & 480 \\
 EfficientNet B7 & 600  & 38.4 & 66.3    & 289.9  & 157    & 652 \\
 EfficientNet B6 & 528  & 19.5 & 43.0    & 167.4  & 246    & 849 \\
 ViT G/14            & 224  & 484.2 & 1844.4 & 275.4 & - \superdagger  & 379 \\
 ViT G/14            & 518  & 2826.1 & 1846.3 & 2639.0 & - \superdagger   & 56\\
 ViT H/14            & 224  & 167.5 & 632.0 & 139.4 & 246    & 960 \\
 ViT H/14            & 392  & 545.9 & 632.7 & 638.0 & 56  & 242 \\
 ViT H/14            & 518  & 1018.8 & 633.5 & 1523.9 & 19    & 116\\
 ViT L/16            & 224  & 61.7 & 304.3 & 63.5 & 701    & 2092 \\
 ViT L/16            & 384  & 191.5 & 304.7 & 270.2 & 177    & 567\\
 ViT L/16            & 512  & 362.9 & 305.2 & 656.4 & 70    & 255\\
 ViT B/16            & 224  & 17.6 & 86.6 & 23.9 & 2247    & 3861 \\
 ViT B/16            & 384  & 55.6 & 86.9 & 101.6 & 549    & 1161 \\
 ViT L/32            & 224  & 15.4 & 306.5 & 13.3 & 3176    & 4431 \\
 ViT L/32            & 384  & 54.4 & 306.6 & 43.9 & 921    & 1439 \\
 RegNetY 128GF & 224 & 127.7 & 644.8 & 71.6  & 191    & 879 \\
 RegNetY 128GF & 384 & 375.2 & 644.8 & 210.2 & 69    & 307 \\
 RegNetY 32GF & 224 & 32.6 & 145.0 & 30.3  & 607 & 2824 \\
 RegNetY 32GF & 384 & 95.1 & 145.0 & 88.9 & 248 & 976 \\
 RegNetY 16GF & 224 & 16.0 & 83.6 & 23.0  & 989 & 4562 \\
 RegNetY 16GF & 384 & 47.0 & 83.6 & 67.7 & 440 & 1401 \\

\end{tabular}}\vspace{1.5mm}
\caption{\textbf{Model complexity and speed.} Complexity and speed of models with an ImageNet-1k head at relevant resolutions. We measure train and train and test speed on a single node with 8 V100 32GB GPUs, maximizing the batch size for each model. Although EfficientNets have very few FLOPs, they produce a large amount of activations resulting in much slower train / test speeds. Training speeds measured for convolutional networks using SGD, and for ViTs using AdamW \cite{loshchilov2017decoupled}. \superdagger We were unable to train a \vitarch{G/14} using our setup, even with a batch size of 1.} 
\label{tab:model_complexity}
\end{table}

\section{Model and Hyperparameter Selection}

\subsection{Effect of Convolutional Model Family}
\label{app:conv_model_family}

We performed experiments investigating the performance of four different model families in weakly-supervised pre-training: \resnext \cite{xie2017aggregated}, \regnety \cite{radosavovic2020designing}, \densenet \cite{huang2017densely} and \efficientnet \cite{tan2019efficientnet}.
As recent model families like \efficientnet and \regnety use squeeze-and-excitation (SE) layers \cite{hu2018squeeze} for improved accuracies \cite{hu2018squeeze, radosavovic2020designing}, we also use these in our implementations of \densenet and \resnext.

\begin{table}[t]\centering
\resizebox{\columnwidth}{!}{\tablestyle{4pt}{1.05}
\begin{tabular}{@{}lc|ccc|cc|ccc@{}}
\bf Model & \bf Res.    & \bf FLOPs  & \bf Param. & \bf Act. & \multicolumn{2}{c|}{\bf Throughput} & \multicolumn{3}{c}{\bf Classification accuracy}\\
           &                       & (B)      &    (M)   &  (M)  & Train & Test & IN-1k & IN-5k & \igdataset[0.7B] \\
           &                       &      &       &    &  &  & &  & $\rightarrow$ IN-1k\\\midrule
 ResNeXt-101 32x4d & $224$  & 8.0 & 49.0 & 21.3 & 2,222 & 5,214 & \underline{79.1} & \underline{50.9} & \underline{80.0} \\
 DenseNet-264 & $224$ &  5.9 & 33.4 & 8.5 & 1,813 & 5,116 & 76.6 & 47.9 & 78.4 \\
 EfficientNet B3 & $300$  & 1.9 & 12.2 & 23.8 & 1,802 & 2,979 & 78.5 & 49.3 & 77.9 \\
 RegNetY 8GF & $224$ & 8.0 & 39.2 & 18.0  & 1,770 & 4,562 & \textbf{79.8} & \textbf{51.4} & \textbf{80.8}\\
\end{tabular}}
\vspace{2mm}
\caption{Overview of the convolutional models we evaluated for our experiments. ResNeXt and DenseNet models were augmented with squeeze-and-excitation (SE~\cite{hu2018squeeze}) layers. %
We evaluate the classification accuracy of the models in three settings: (1) training on \indataset{1}; (2) training on \indataset{5}; and (3) pre-training 1 epoch on 1B examples of \igdataset[0.7B] followed by linear classifier evaluation on \indataset{1}. We find that the RegNetY model performs best in all settings. The best result on each dataset is \textbf{boldfaced}; the second-best result is \underline{underlined}. Higher is better.}
\label{tab:candidate_models}
\end{table}

Since we trained our models at scale, our goal was to identify the most efficient model family in terms of the accuracy achieved with a fixed training budget. 
In line with this goal, instead of finding models with comparable numbers of FLOPs or parameters, which have been shown to correlate poorly with training speed~\cite{dollar2021fast,radosavovic2020designing}, we instead measured image throughput during training. 
We also include the test time throughput as well since it is a useful inference time constraint to consider.

To keep the experiments tractable, we used medium-sized models of each model family. 
Table~\ref{tab:candidate_models} lists the candidate models for each of the families we used for our comparison; these models were selected to have similar training speeds (in terms of images processed per second). We note that the test throughputs were also similar except for the \efficientnet model which uses a higher resolution than the other models.

We tested the models in three settings: (1) training and testing on \indataset{1}, (2) training and testing on \indataset{5}, and (3) pre-training on \igdataset[1B] followed by a linear classifier trained and tested on \indataset{1}.
The results of our experiments are presented in Table~\ref{tab:candidate_models}. 
The results show that for a similar training budget, the \regnety model family outperforms the other model families on all three datasets, while also having a competitive inference throughput.
For that reason, we focused on \regnety models in all subsequent experiments.

\subsection{Effect of Dataset Size}
\label{app:dataset_size}

During pre-training, usually the focus is on the the total number of unique images in the dataset, which we will refer to as the dataset \emph{size} \cite{mahajan2018exploring, xie2020self, dosovitskiy2020image, zhai2021scaling}.
In our setup, due to the upsampling of the least frequent hashtags, our final dataset is defined by an additional parameter -- the dataset's \emph{samples}, which we define as the total number of image-label pairs, counting duplicates.
Table \ref{tab:dataset_size} shows the effect of a dataset's number of unique images vs the total samples seen during training.
For the IG dataset in the smaller test regimes we explored, the total samples seen determined the model performance across a variety of dataset sizes for different model families (convolutional networks, transformers) and model capacities, rather than the number of unique images seen. We hypothesize that this is because in this regime the model has not yet saturated. It does suggest that the total number of samples seen during training is important to consider when comparing large datasets with a small number of epochs.

\begin{table}[t]\centering
\resizebox{0.7\columnwidth}{!}{
\tablestyle{4pt}{1}
\begin{tabular}{@{}cc|c|cc|cc@{}}
 \multicolumn{2}{c|}{\bf Dataset} & \bf Epochs &  \multicolumn{4}{c}{\bf IN-1k transfer accuracy} \\
\bf Name & \bf Samples &               & \multicolumn{2}{c|}{\bf RegNetY} & \multicolumn{2}{c}{\bf ViT} \\
\bf (size) & \bf                    &               & \bf 8GF & \bf 32GF & \bf B/16 & \bf L/16 \\\midrule
 \igdataset[0.2B] & 250M  & 8               & 81.5 & 83.7 & 80.5 & 83.2 \\
 \igdataset[0.4B] & 500M & 4               & 81.5 & 83.8 & 80.7 & 83.5 \\
 \igdataset[0.7B] & 1B & 2               & 81.4 & 83.8 & 80.3 & 83.5 \\
 \igdataset[1.4B] & 2B  & 1               & 81.3 & 83.7 & 80.5 & 83.4 \\
\end{tabular}
}
\vspace{2mm}
\caption{\textbf{Effect of dataset size}. We compute the linear classifier accuracy of various models on \indataset{1} to study the effect of unique images. Every data point corresponds to the same number of total samples trained (2 billion), but the dataset size varies.}  
\label{tab:dataset_size}
\end{table}

\subsection{Effect of Scaling Parameters}
Due to the inherent noise in the learning signal, weakly supervised pre-training requires substantial scale to obtain optimal results.
We performed experiments studying the effect on the transfer performance of two key scaling parameters: (1) model scale and (2) training set scale.
To vary the model scale, we train \regnety models that were independently optimized for use, starting from 16 GFLOPs, up to 128 GFLOPs.
We followed \cite{radosavovic2020designing} and searched for each of the models instances on \indataset{1}.
To vary the training set scale, we use IG datasets of varying sizes.
We train all models for one full epoch, and measure linear classifier performance on \indataset{1}.

The results of our experiments are presented in Figure~\ref{fig:scaling}.
We present the transfer accuracy as a function of both the total samples seen and the total training time in GPU-days, for four different models.
The results presented in Figure~\ref{fig:scaling} are largely in line with those of~\cite{mahajan2018exploring,zhai2021scaling}.
Specifically, transfer accuracy improves for both larger models and for longer training regimes.
Akin to~\cite{mahajan2018exploring}, we find that the larger models benefit from more training samples than their smaller counterparts: the slope of the accuracy curve of \regnetyarch{128} is steeper than that of \regnetyarch{16}.
Thus, for a large enough training budget it makes sense to use a larger model rather than a smaller model trained on more samples.

\begin{figure}[t]\centering
\includegraphics[width=\columnwidth]{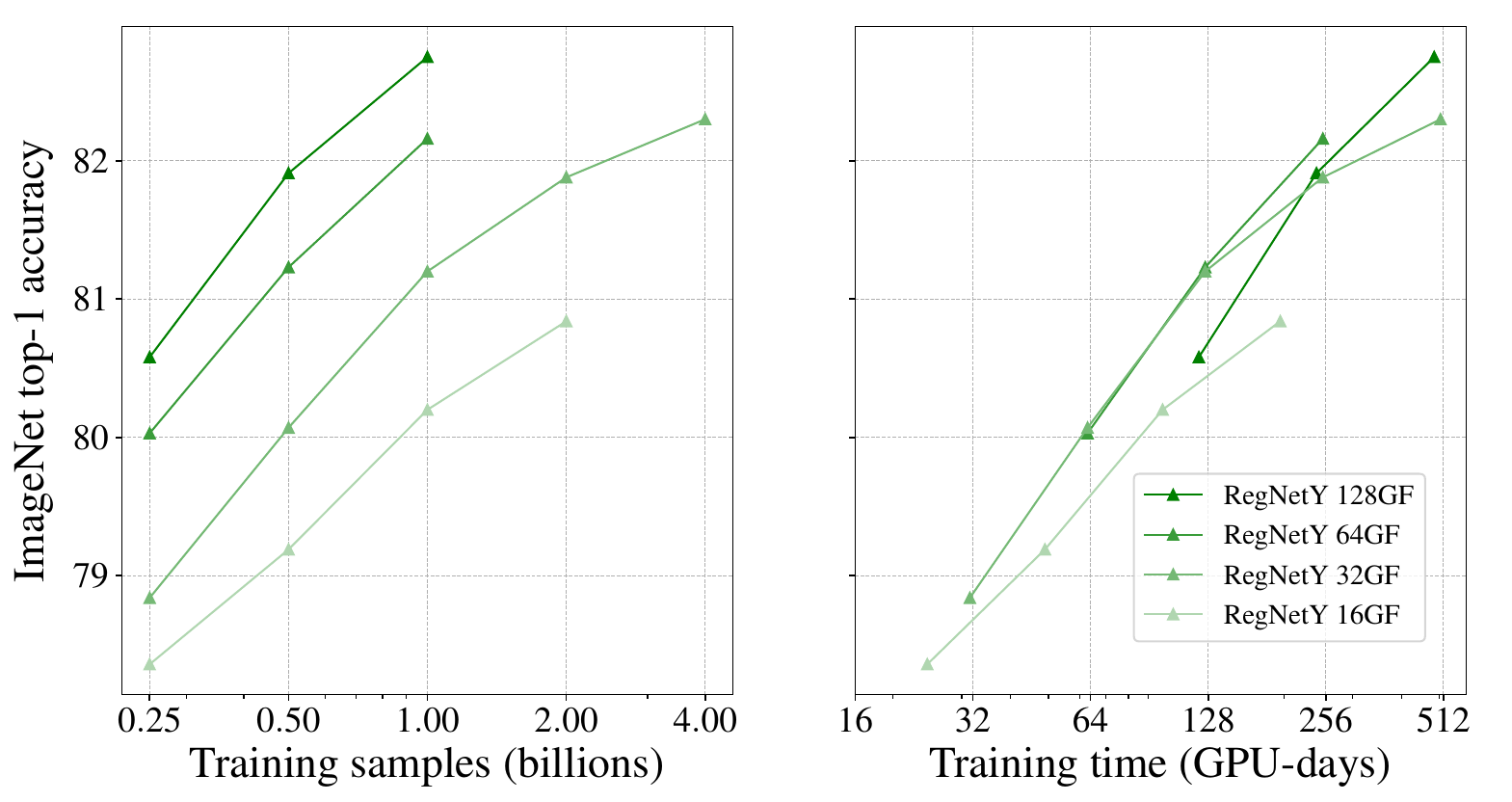}
\caption{\textbf{Scaling model and dataset sizes.} ImageNet top-1 linear classifier accuracy for various model sizes as a function of the number of pre-training samples (\textbf{left}) and the training budget (\textbf{right}). As we go larger in model size, the models become more efficient in utilizing a given number of pre-training samples, and additional training samples improve performance. %
Training time calculated by dividing the total samples with the training speeds from Table \ref{tab:model_complexity}.}
\label{fig:scaling}
\end{figure}

\section{Training Details}
All our models were trained with \emph{Classy Vision} \cite{adcock2019classy}. For the transfer results for other works\iftoggle{cvprfinal}{ in Table \ref{tab:transfer_results}}{}, we used the \emph{timm} library \cite{rw2019timm} to get pre-trained checkpoints. In this section we share details about our fine-tuning setup\iftoggle{cvprfinal}{ for Table \ref{tab:transfer_results}}{}, viz.{,} the learning rate used (Table \ref{tab:transfer_lr}), and the utility of using synchronized batch normalization for convolutional networks (Figure \ref{fig:sync_bn}).

\paragraph{Hashtag-to-class mapping in zero-shot experiments.} 
Because both the ImageNet and \igdataset datasets have target sets drawn from English nouns, we can construct a many-to-many mapping from Instagram hashtags to ImageNet classes.
We first map each hashtag to all WordNet synsets of the hashtag, and then map each ImageNet class to the most similar hashtag(s) based on the maximum path similarity score in WordNet \cite{fellbaum1998wordnet} between any of the the hashtag synsets and the ImageNet class synset.
As the hashtags are nouns or compound nouns, they can have multiple meanings: for example, \texttt{\#crane} refers to both the bird and the building structure.
However, the synset of \texttt{crane} referring to the bird and synset of \texttt{crane} referring to the structure are two distinct ImageNet classes.
In this situation, we map both synsets to \texttt{\#crane}. %
Likewise, a synset can represent a concept specified by multiple words and therefore by multiple hashtags, for example, the synset \texttt{\{porcupine, hedgehog\}} matches both \texttt{\#porcupine} and \texttt{\#hedgehog}.
In this case, we map the synset to all corresponding hashtags.

\begin{figure}[t]\centering
\includegraphics[width=\columnwidth]{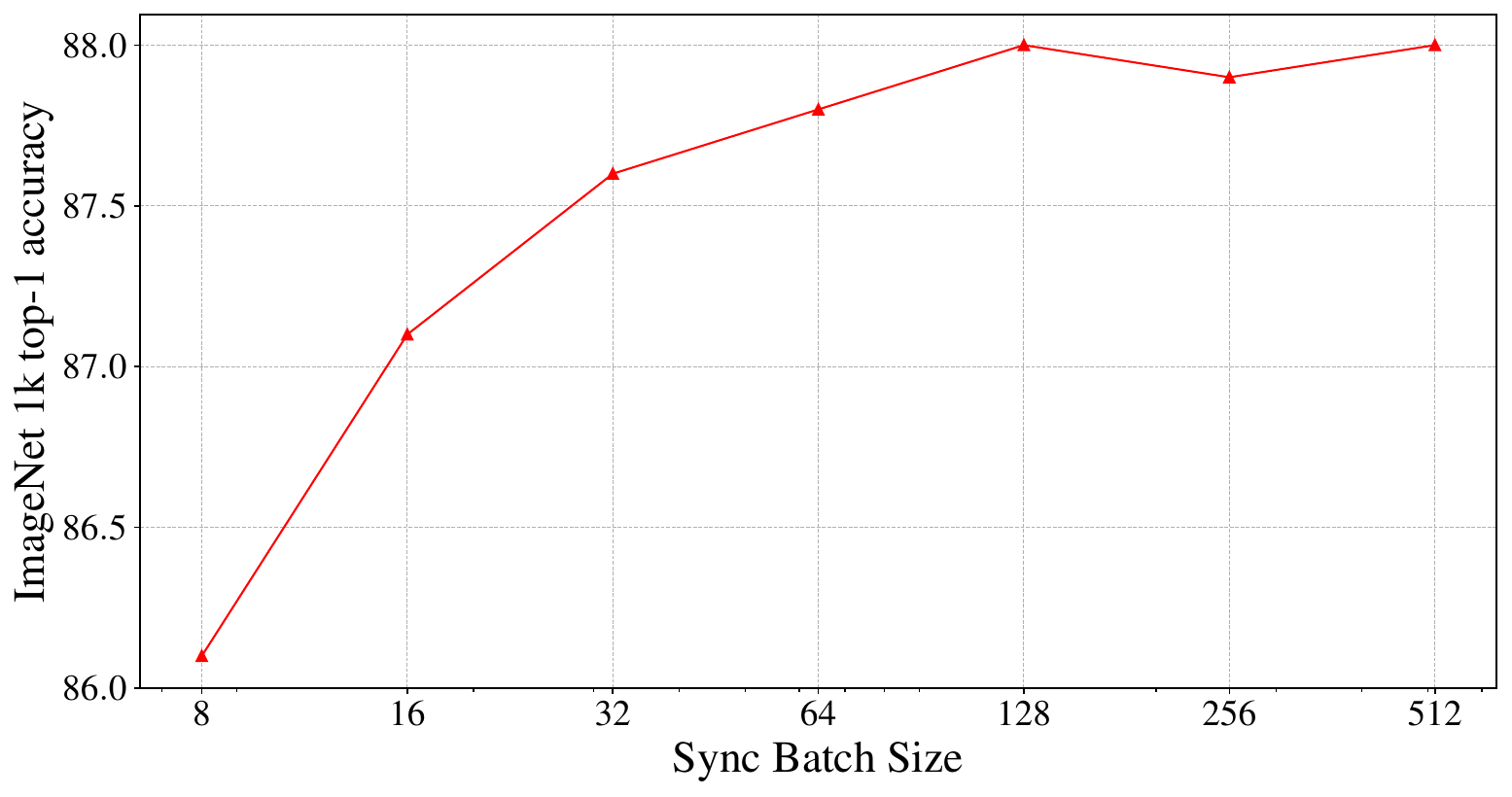}
\caption{\textbf{Effect of sync batch norm while fine-tuning.} ImageNet top-1 accuracy while fine tuning a \regnetyarch{128} continues to increase as we increase the sync batch size. The model was always trained with a mini batch size of $512$, while varying the batch sizes for sync batch norm. Results reported without EMA. %
}
\label{fig:sync_bn}
\end{figure}

To utilize the resulting many-to-many mapping between hashtags and ImageNet classes, we need to aggregate the model (hashtag) predictions into predictions over the ImageNet classes.
For the RegNetY models, we first map the prediction value for a hashtag to all ImageNet classes that the hashtag maps to.
When Platt scaling is used, we \emph{sum} all the resulting values for an ImageNet class to aggregate them. 
When Platt scaling is not used, we instead \emph{average} the predicted values for a class.
For the ViT models, we achieved better results with a different aggregation method:
we map $\nicefrac{1}{N}$ of the prediction value for a hashtag to all $N$ ImageNet classes that the hashtag maps to, and take the maximum over all the resulting values for each class.

\begin{table}[t]\centering
\resizebox{\columnwidth}{!}{\tablestyle{4pt}{1}
\begin{tabular}{@{}lc|ccccc@{}}
\bf Model  & \bf Pre-training  &  \multicolumn{5}{c}{\bf Learning rate} \\
               & & \bf IN-1k & \bf IN-5k & \bf iNat & \bf Places & \bf CUB  \\\midrule
 \multicolumn{6}{l}{\textit{Supervised pre-training}}\\\midrule
 EfficientNet L2~\cite{xie2020self} & JFT 300M & -- & \num{2.0e-1} & \num{2.0e-1} & \num{9.6e-2}  & \num{2.0e-1} \\
 EfficientNet B7~\cite{xie2020self} & JFT 300M & -- & \num{2.0e-1} & \num{2.0e-1} & \num{2.0e-1} & \num{2.0e-1} \\
 EfficientNet B6~\cite{xie2020self} & JFT 300M & -- & \num{2.0e-1} & \num{2.0e-1} & \num{2.0e-1} & \num{2.0e-1} \\
 EfficientNet B8~\cite{xie2020adversarial} & IN-1k & -- & \num{9.6e-2} & \num{4.0e-1} & \num{9.6e-2} & \num{2.0e-1} \\
 EfficientNet B7~\cite{xie2020adversarial} & IN-1k & -- & \num{9.6e-2} & \num{4.0e-1} & \num{9.6e-2} & \num{2.0e-1} \\
 EfficientNet B6~\cite{xie2020adversarial} & IN-1k & -- & \num{9.6e-2} & \num{4.0e-1} & \num{9.6e-2} & \num{2.0e-1} \\
 ViT L/16~\cite{dosovitskiy2020image} & IN-21k & -- & -- & \num{4.8e-2} & \num{2.4e-2}  & \num{4.8e-2} \\
 ViT B/16~\cite{dosovitskiy2020image} & IN-21k & -- & -- & \num{4.8e-2} & \num{2.4e-2}  & \num{4.8e-2} \\
 ViT L/32~\cite{dosovitskiy2020image} & IN-21k & -- & -- & \num{4.8e-2} & \num{2.4e-2}  & \num{4.8e-2} \\\midrule
 \multicolumn{6}{l}{\textit{Weakly supervised pre-training}}\\\midrule
 ViT H/14 & IG 3.6B & \num{6.0e-3} &\num{2.4e-2} & \num{1.2e-2} & \num{3.0e-3}  & \num{6.0e-3} \\
 ViT L/16 & IG 3.6B &\num{6.0e-3} &\num{2.4e-2} & \num{1.2e-2} & \num{3.0e-3} & \num{3.0e-3}  \\
 ViT B/16 & IG 3.6B &\num{6.0e-3} &\num{2.4e-2} & \num{1.2e-2} & \num{3.0e-3} & \num{3.0e-3} \\
 RegNetY 128GF & IG 3.6B &\num{6.0e-3} &\num{2.4e-2} & \num{1.2e-2} & \num{6.0e-3} & \num{3.0e-3}  \\
 RegNetY 32GF & IG 3.6B &\num{6.0e-3} & \num{1.2e-2} & \num{1.2e-2} & \num{1.2e-2} & \num{6.0e-3}  \\
 RegNetY 16GF & IG 3.6B &\num{6.0e-3} & \num{1.2e-2} & \num{1.2e-2} & \num{1.2e-2} & \num{6.0e-3}  \\
\end{tabular}}
\vspace{2mm}
\caption{Base learning rate used for the transfer results\iftoggle{cvprfinal}{ in Table \ref{tab:transfer_results}}{}. } 
\label{tab:transfer_lr}
\end{table}

\section{ImageNet Robustness Experiments}
A potential advantage of weakly supervised pre-training is that the resulting models have observed more training images.
This may lead the model to be more robust to variations in the image content. To evaluate the robustness of our models under small variations in visual content, image distribution, or labeling, we performed additional transfer-learning experiments using three ImageNet-like datasets: (1) ReaL ImageNet~\cite{beyer2020imagenet}, (2) ImageNet v2~\cite{recht2019imagenet}, and (3) ObjectNet~\cite{barbu2019objectnet}. %
We fine-tune pre-trained models on the ImageNet-1k dataset and test them directly on the three evaluation datasets.

Table~\ref{tab:baselines_in} presents the results of this experiment.
While the highest accuracies are obtained by large vision transformers (ViT) trained on 3 billion labeled images (JFT 3B), our weakly pre-trained RegNetY and ViT models are very competitive: our largest models are the runner-up on each of the datasets.
In terms of differences in robustness, however, the results are inconclusive: validation accuracy on ImageNet-1k appears to be a good predictor for accuracy on the other tests sets across models and training regimes.

\begin{table}[t]\centering
\resizebox{0.9\columnwidth}{!}{
\tablestyle{4pt}{1}
\begin{tabular}{@{}lc|cccc@{}}
\bf Model  & \bf Pre-training  &  \multicolumn{4}{c}{\bf Classification accuracy} \\
               & & \bf IN-1k & \bf ReaL-IN & \bf IN-v2 & \bf Obj. Net  \\\midrule
 \multicolumn{6}{l}{\textit{Supervised pre-training}$^\dagger$}\\\midrule
 EfficientNet L2~\cite{xie2020self} & JFT 300M$^\ddagger$ & \emph{88.4} & 90.6 & 80.2 & 68.3 \\
 EfficientNet B7~\cite{xie2020self} & JFT 300M$^\ddagger$ & \emph{86.9} & 90.1 & 78.1 & 61.2\\
 EfficientNet B6~\cite{xie2020self} & JFT 300M$^\ddagger$ & \emph{86.4} & 89.8 & 76.7 & 60.0\\
 EfficientNet B8~\cite{xie2020adversarial} & IN-1k & \emph{85.5} & 89.6 & 76.1 & 54.5\\
 EfficientNet B7~\cite{xie2020adversarial} & IN-1k & \emph{85.2} & 89.5 & 75.7 & 53.3\\
 EfficientNet B6~\cite{xie2020adversarial} & IN-1k & \emph{84.8} & 89.4 & 75.5 & 51.9\\
 ViT G/14~\cite{zhai2021scaling} & JFT 3B & \textbf{\emph{90.5}} & \textbf{\emph{90.8}} & \textbf{\emph{83.3}} & \textbf{\emph{70.5}}\\
 ViT L/16~\cite{zhai2021scaling} & JFT 3B & \emph{88.5} & \emph{90.4} & \emph{80.4} & --\\
 ViT H/14~\cite{dosovitskiy2020image} & JFT 300M & \underline{\emph{88.6}} & \underline{\emph{90.7}} & -- & --\\
 ViT L/16~\cite{dosovitskiy2020image} & JFT 300M  & \emph{87.8} & \emph{90.5} & -- & -- \\
 ViT H/14~\cite{dosovitskiy2020image} & IN-21k & \emph{85.1} & \emph{88.7} & -- & -- \\
 ViT L/16~\cite{dosovitskiy2020image} & IN-21k & 85.2 & \emph{88.4} & 74.8 & 56.6 \\
 ViT B/16~\cite{dosovitskiy2020image} & IN-21k & 84.2 & \emph{88.4} & 73.5 & 52.6 \\
 ViT L/32~\cite{dosovitskiy2020image} & IN-21k & 81.5 & \emph{86.6} & 71.2 & 47.2 \\\midrule
 \multicolumn{6}{l}{\textit{Weakly supervised pre-training}}\\\midrule
 ViT H/14 & IG 3.6B & \underline{88.6} & 90.5 & \underline{81.1} & \underline{69.5} \\
 ViT L/16 & IG 3.6B & 88.1 & 90.6 & 80.3 & 66.2 \\
 ViT B/16 & IG 3.6B & 85.3 & 89.1 & 75.6 & 55.2 \\
 RegNetY 128GF & IG 3.6B & 88.2 & \underline{90.7} & 80.4 & 68.5 \\
 RegNetY 32GF & IG 3.6B & 86.8 & 90.2 & 78.2 & 62.5 \\
 RegNetY 16GF & IG 3.6B & 86.0 & 89.9 & 76.9 & 59.0 \\
\end{tabular}}
\vspace{2mm}
\caption{Classification accuracy of models pre-trained on the specified pre-training dataset followed by finetuning on ImageNet-1k. Accuracy is measured on four ImageNet-like datasets: (1) ImageNet-1k itself, (2) ReaL ImageNet~\cite{beyer2020imagenet}, (3) ImageNet v2~\cite{recht2019imagenet}, and (4) ObjectNet~\cite{barbu2019objectnet}.  The best result on each dataset is \textbf{boldfaced}; the second-best result is \underline{underlined}. Numbers that are adopted from the original paper are \emph{italicized}. Higher is better. $^\dagger$It is unknown how much manual curation was performed in the annotation of JFT datasets. $^\ddagger$Pre-training data also includes IN-1k.} 
\label{tab:baselines_in}
\end{table}

\section{Broader Impact}
\label{sec:broader_impact}
This section presents a more detailed account of the experiments presented in the main paper, which aim to understand: (1) how well our models perform on photos taken in non-English speaking countries, and (2) the associations our hashtag-prediction models learn with photos of people with varying characteristics. In this section we share and discuss all the experimental results in more detail. 
As a reminder, all results presented below are for the hashtag-prediction models; no fine-tuning is employed.

\subsection{Analyzing Hashtag Prediction Fairness}
\label{sec:fairness_analysis}
Following prior work~\cite{devries2019}, we analyzed how well the RegNetY 128GF model works on photos taken across the world. We first repeated the analysis of~\cite{devries2019} on the Dollar Street dataset. 
To this end, we use the hashtag-prediction model in a zero-shot fashion: we manually define a mapping from hashtags to the 112 classes in the Dollar Street, and task the model with predicting only hashtags that are mapped to a class. 
We measure the accuracy of the model’s predictions per country, and display the results on a world map in which colors correspond to accuracies in the left plot in Figure~\ref{fig:world_map} (red is $40\%$ correct; green is $70\%$).

\begin{figure}[h]
\centering
\begin{tabular}{c}
\includegraphics[width=\linewidth]{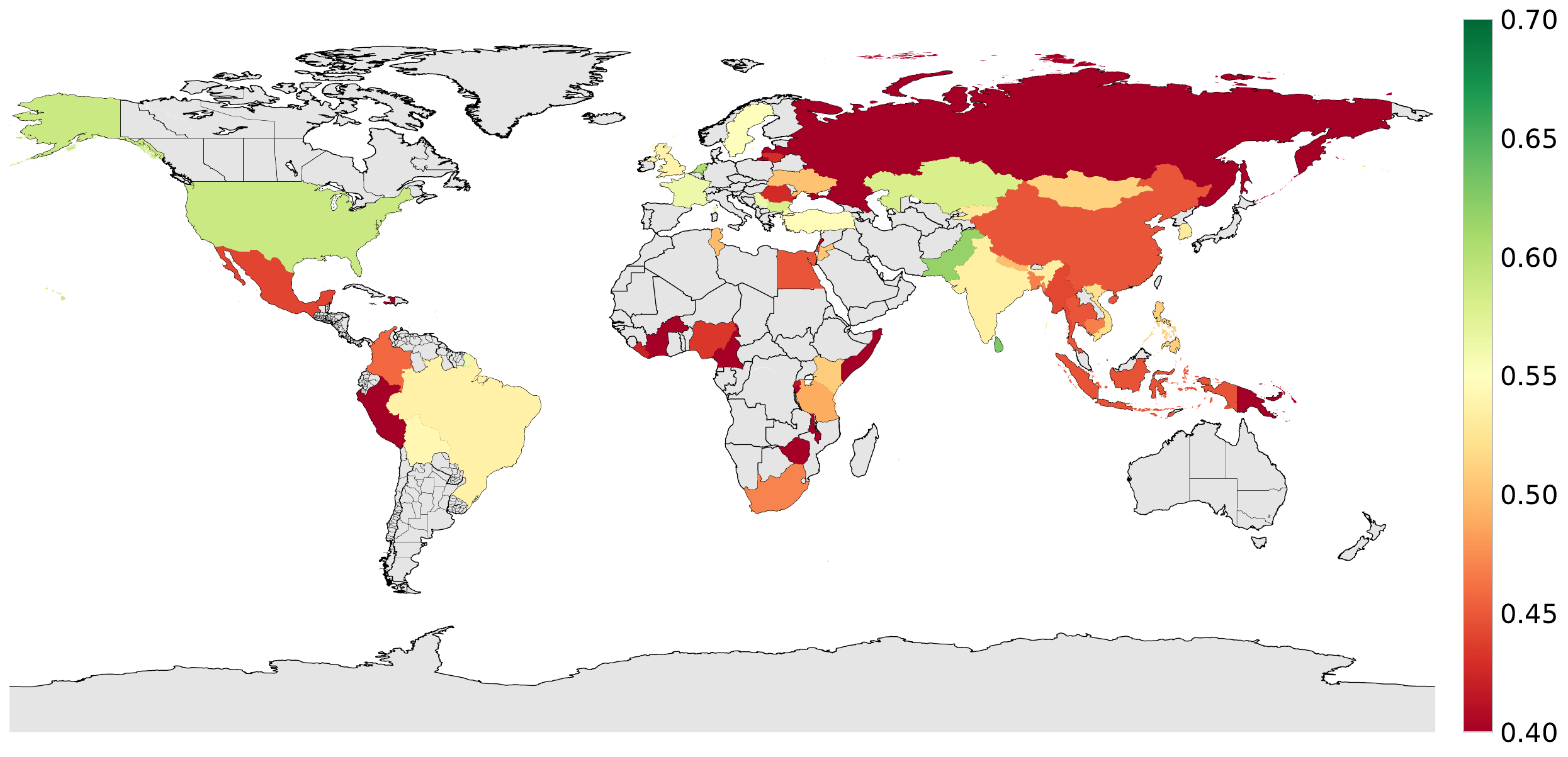} \\
\includegraphics[width=\linewidth]{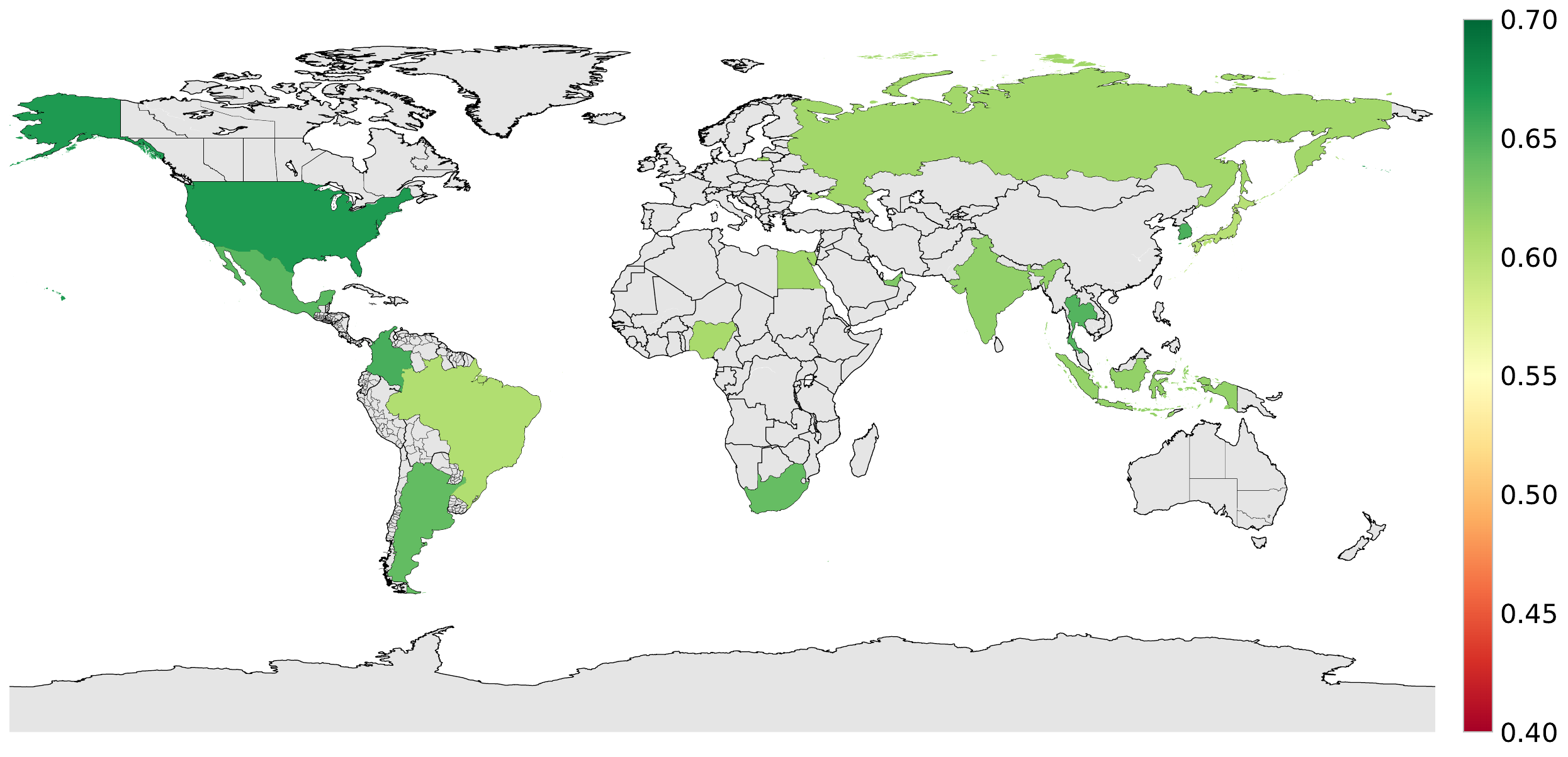}
\end{tabular}
\caption{Recognition accuracy per country of our zero-shot classifier on the Dollar Street dataset (\textbf{top}) and a proprietary dataset (\textbf{bottom}). The accuracy on all images is $48.0\%$ on the Dollar Street dataset and $63.3\%$ on the proprietary dataset.}
\label{fig:world_map}
\end{figure}

Although the absolute numbers are lower because the image-recognition model operates in zero-shot mode (the average accuracy over all countries is $48.0\%$), qualitatively, the observations we obtain are in line with prior work~\cite{devries2019}: observed recognition accuracies are higher in the US and Europe than in most other countries.

Because the Dollar Street dataset may itself have issues, we repeated the analysis on a proprietary dataset that contains millions of images labeled for visual concepts and their country of origin. 
The resulting world map is shown in the right plot in Figure~\ref{fig:world_map}.
The results suggest that the range of accuracy values is relatively tight (approximately $5\%$) on this large proprietary dataset.

Following common practice~\cite{zafar2017fairness}, we also measure the percentage of classes for which the ratio between the class-recognition accuracy in country 1 and country 2 is smaller than 0.8. 
The results of this analysis are shown in the heat map in Figure~\ref{fig:internal_country_differences_results}.
If an entry in the heat map is yellow, then the model recognizes a substantial percentage (up to $35\%$) of classes substantially worse in the ``row country'' than in the ``column country''.

\begin{figure}[t]
\centering
\includegraphics[width=0.8\linewidth]{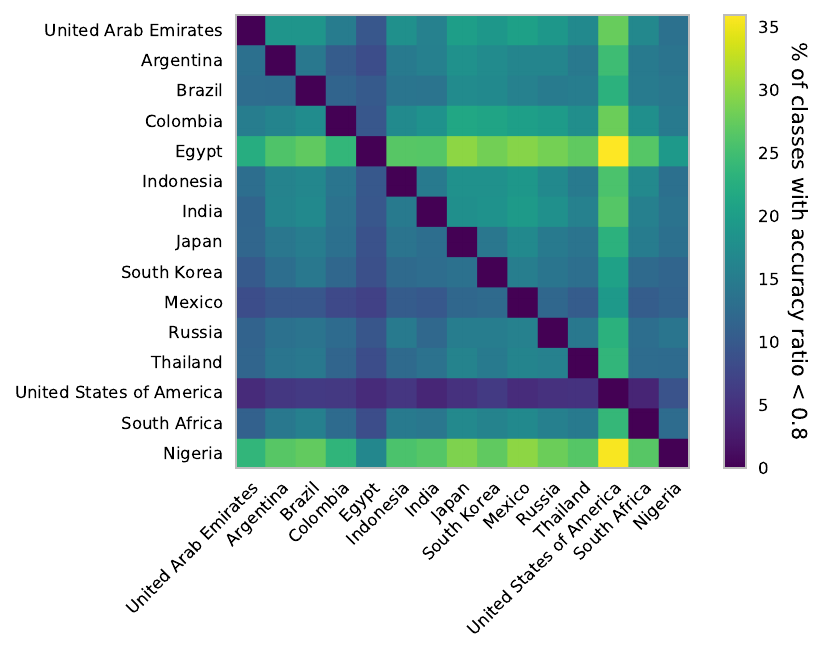}
\caption{Percentage of classes for which recognition accuracy is substantially higher in one country (\textbf{rows}) than in another country (\textbf{columns}). We use the $80\%$ rule to assess whether one accuracy is ``substantially higher'' than the other.}
\label{fig:internal_country_differences_results}
\end{figure}

The results in the figure suggest that the hashtag-prediction model performs better in the US and worse in Egypt and Nigeria. 
There is also a notable difference between the accuracy map in Figure~\ref{fig:world_map} and the heat map in Figure~\ref{fig:internal_country_differences_results}. 
The accuracy map suggests that the hashtag-prediction model performs worst in Brazil and Japan, whereas the heat map suggests the lowest accuracy is obtained in Egypt and Nigeria. 
This result may be due to variations between countries in the distribution of per-class accuracy discrepancies and/or due to variations in the concept distribution per class.

\subsection{Analyzing Associations in Hashtag Predictions}
We performed experiments in which we analyze the associations our hashtag-prediction models make for photos of people with different apparent skin tone, different apparent age, different apparent gender, and different apparent race.
We present the results of each experiment separately below.

\paragraph{Apparent Skin Tone}
\label{sec:skintone_analysis}
We first evaluated potentially troubling associations in hashtag predictions by apparent skin tone. 
To this end, we used a proprietary dataset that contains $178{,}448$ Instagram photos that were annotated using the Fitzpatrick skin tone scale~\cite{fitzpatrick1975}. 
We ran all these photos through our RegNetY 128GF hashtag-prediction model, asking it to predict the five highest-scoring hashtags for each photo. 
We maintain per-skin tone statistics on how often a hashtag in the vocabulary is predicted for a photo of an individual with that skin tone. 
Next, we inspect differences in the hashtag prediction rate between different skin tones. 
For each skin tone, we identify the hashtags with the largest absolute difference in hashtag prediction rate compared to the average prediction rate for the other five skin tones. We also compute the associated relative difference in hashtag prediction rate. 
We show the resulting hashtags for skin tone 1 (lightest skin tone) and skin tone 6 (darkest skin tone) in the top row of Figure~\ref{fig:rai_results}.

\begin{figure*}[t]
\centering
\includegraphics[width=14.4cm]{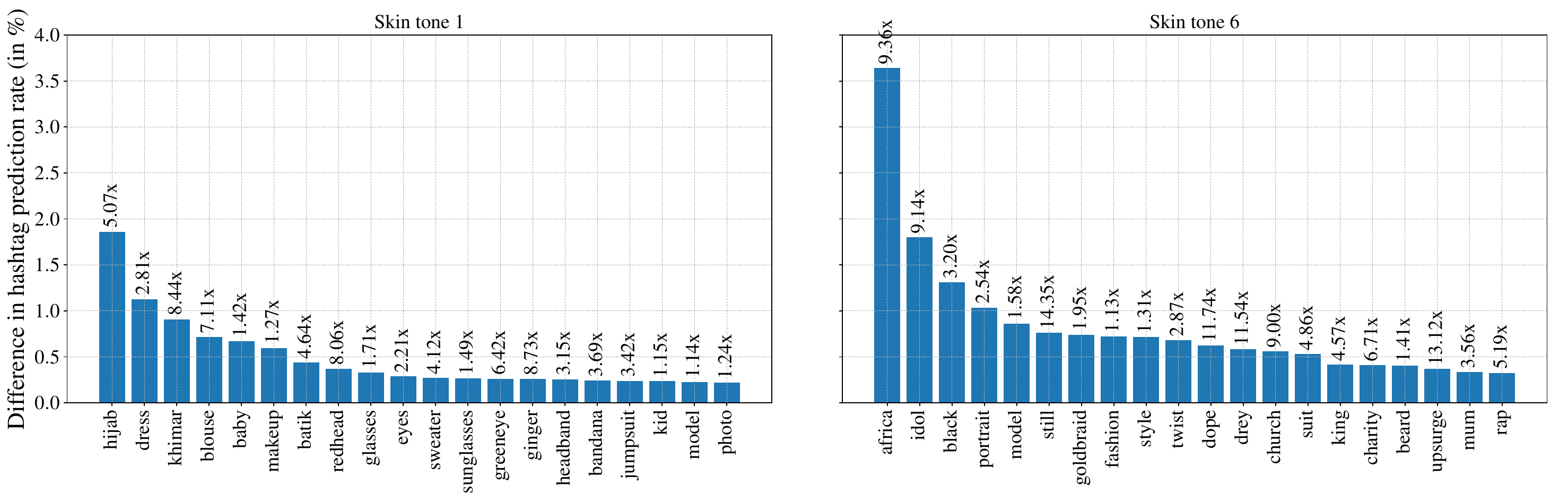}
\includegraphics[width=14.4cm]{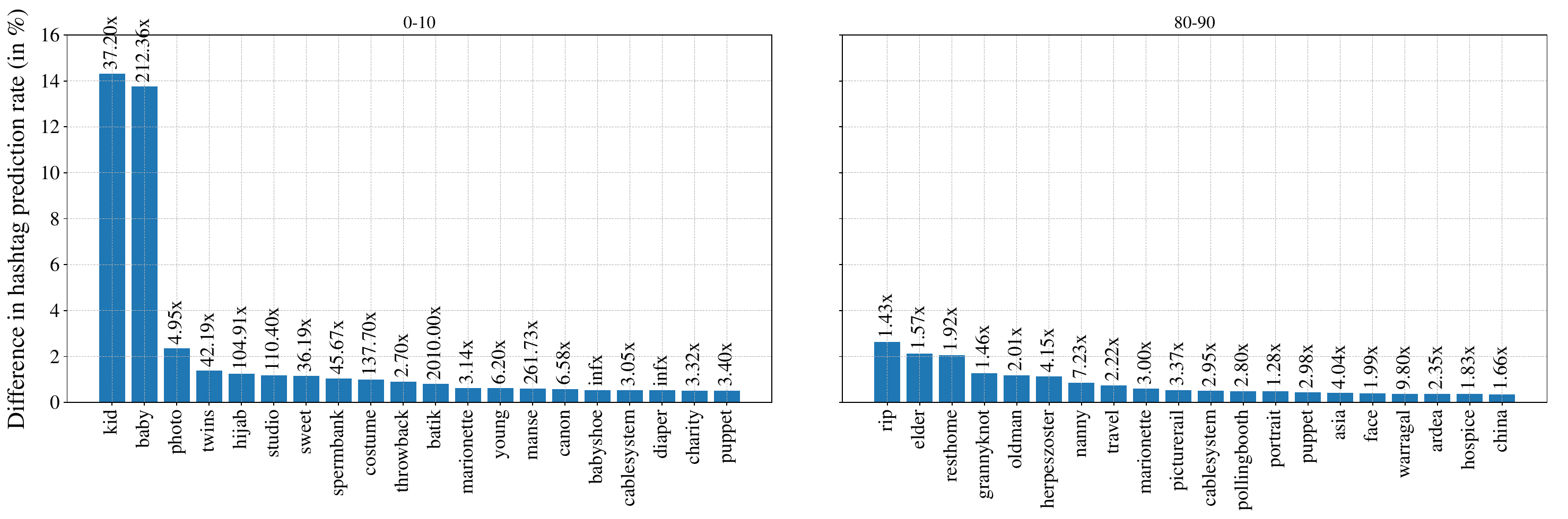}
\includegraphics[width=14.4cm]{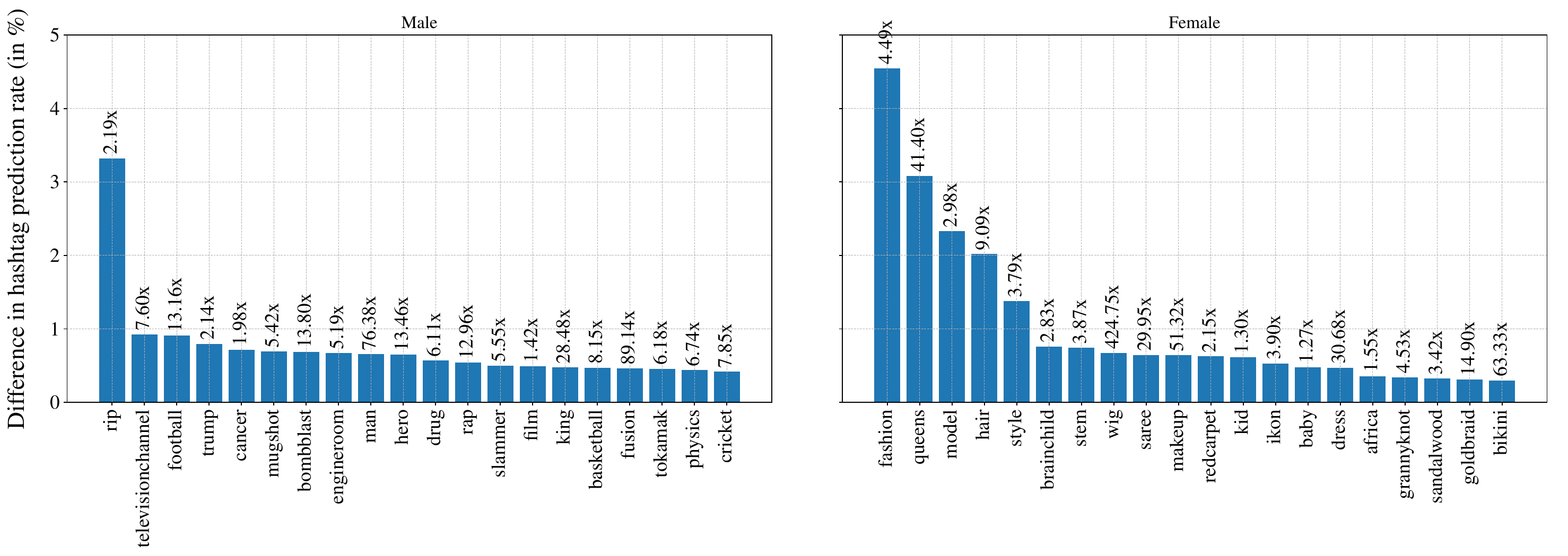}
\includegraphics[width=14.4cm]{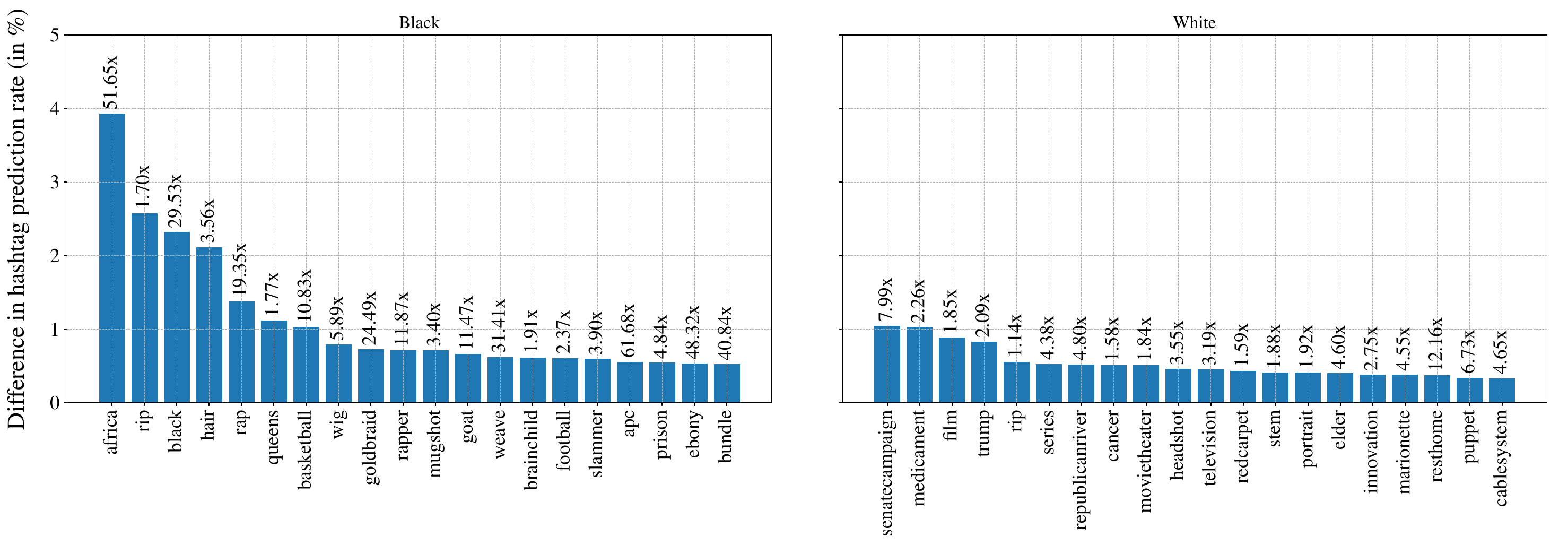}
\caption{Differences in hashtag prediction rate for photos from various apparent subgroups. Absolute differences are sorted, and results for 20 hashtags with the largest difference are shown. Relative hashtag prediction differences are shown on top of the bars. From \textbf{top} to \textbf{bottom}: Differences for photos of people with (apparent) Fitzpatrick skin tone 1 and photos of people with other apparent skin tones (\textbf{left}); and between photos with skin tone 6 and other skin tones (\textbf{right}). Differences between photos with (apparent) age group 0-10 and other age groups; and between age group 80-90 and other age groups. Differences between photos of (apparent) women and photos of men; and between photos of men and women. Differences between photos of (apparent) Black people and people of other races; and between photos of White people and other races.}
\label{fig:rai_results}
\end{figure*}

The results in the figure reveal several associations that may not be unexpected: for example, \texttt{\#redhead} is more commonly predicted by the model for photos of people with a light skin tone, whereas \texttt{\#black} is more often predicted for people with a dark skin tone. 
The analysis also reveals associations that are more difficult to explain: do people with lighter skin tones wear more \texttt{\#headbands} or \texttt{\#bandanas}? 
It is also unclear to what extent the associations we find are learned by the model and to what extent they reflect characteristics of the evaluation data.

\paragraph{Apparent Age}
\label{sec:age_analysis}
We performed a similar analysis of associations between predicted hashtags and apparent age groups. 
For this evaluation, we used the UTK Faces dataset~\cite{zhifei2017utk}, which provides apparent age labels. 
People were grouped into age buckets with a range of $10$ years ($0-10$, $10-20$, $20-30$ years, \emph{etc.}).
We performed the same analysis as before. 
The second row of Figure~\ref{fig:rai_results} shows the most common hashtag predictions for two different (apparent) age groups.

Some associations that the analysis reveals are not unexpected: for example, predicting \texttt{\#baby} or \texttt{\#kid} for age group $1-10$ years or predicting \texttt{\#elder} for age group $80-90$ years. 
The results also show that there may be discrepancies in the meaning of words and hashtags: \texttt{\#rip} is in the hashtag dictionary because one may have a rip in their shirt but it is commonly used on Instagram as abbreviation for ``rest in peace'', which is more likely to apply to people of age. 
Other disparate associations appear unfortunate, such as the association of \texttt{\#spermbank} with photos of people aged $0-10$ years.

\paragraph{Apparent Gender}
\label{sec:gender_analysis}
We performed the same analysis on the UTK Faces dataset~\cite{zhifei2017utk} by apparent gender. 
Due to limitations of the evaluation dataset, we restricted our analyses to males and females but did not consider non-binary genders. 
The results are presented in the third row of Figure~\ref{fig:rai_results}.

The results suggest that the model has learned certain gender-specific stereotypes, for example, associating men with \texttt{\#football} and \texttt{\#basketball} more frequently or associating women more frequently with \texttt{\#makeup} and \texttt{\#bikini}. 
The associations revealed by the analysis vary in how problematic they are: for example, men may not be excited that they are more frequently associated with \texttt{\#mugshot} -- and in some cases, such an association could be harmful. 
We will return to this example below.

\paragraph{Apparent Race}
\label{sec:race_analysis}
For better or worse (see below), the UTK Faces dataset~\cite{zhifei2017utk} also contains annotations of apparent race. 
We repeated the same hashtag prediction analysis for the groups defined in UTK Faces (Indian, Asian, Black, White, Other) as well.
We present the results of this analysis in the fourth row of Figure~\ref{fig:rai_results}. 

The results analysis suggest a variety of disparate associations, some of which are more problematic than others. 
Likely the most troubling association suggested by the analysis is the association of photos of Black people with \texttt{\#mugshot} and \texttt{\#prison}.
Because of the sensitivity of this type of association, we investigated it more in-depth. 
First, we performed a visual analysis of the photos for which the hashtag-prediction model predicted \texttt{\#mugshot} or \texttt{\#prison} among its top-5 predictions. 
This inspection revealed that a small percentage of the photos in the UTK Faces dataset are, indeed, mug shots. Specifically, some of the images in the dataset appear to have been sourced from \url{http://mugshots.com/}. 
This observation raises an important question: \emph{Are the associations our analyses identify due to associations that the model has learned, due to biases in the evaluation data, or both?} 
This question is difficult to answer without collecting additional annotations.

In this particular case, we decided to re-use the skin tone dataset we used earlier and measure how often \texttt{\#mugshot} is predicted for the images in that dataset. 
While skin tone does not map to race very well, we would expect to observe at least some correlation between \texttt{\#mugshot} prediction and skin tone if the model had learned this association. 
The results were quite the opposite: \texttt{\#mugshot} was predicted 7 times ($0.0078\%$) for images with Fitzpatrick skin tone 1 (lightest skin tone) but only once for skin tone 6 (darkest skin tone; $0.0023\%$).
Combined with our visual inspection, this suggests that the problematic association we observed in the analysis on UTK Faces is most likely to be due to problems in the UTK Faces dataset itself than due to problems in the hashtag-prediction model.
Having said that, we acknowledge that there are many caveats here, and that our experiments are not fully conclusive.

}{}

\end{document}